\documentclass[11pt]{article}

\usepackage{lmodern}
\usepackage[left=1.5cm,right=1.5cm,top=2cm,bottom=2cm]{geometry}\usepackage{authblk}

\usepackage{cite}
\usepackage[pdftex]{graphicx}

\usepackage{xr-hyper}
\usepackage{hyperref}

\usepackage[utf8]{inputenc} 
\usepackage[T1]{fontenc}    
\usepackage{hyperref}       
\hypersetup{colorlinks,linkcolor=blue, citecolor=red}
\usepackage{url}            
\usepackage{booktabs}       
\usepackage{amsfonts}       
\usepackage{nicefrac}       
\usepackage{microtype}      

\usepackage{xcolor}

\usepackage{amsmath}
\usepackage{amssymb}
\usepackage{bm}

\usepackage{subfigure}

\usepackage{titlesec}      
\newcommand{\Cmag}{m}
\newcommand{\Cfdt}{C_\text{TTI}}
\newcommand{\Rfdt}{R_\text{TTI}}

\newcommand{\Cov}{\text{Cov}}
\newcommand{\bs}{\pmb\sigma_0} 
\newcommand{\loss}{\text{\large $\ell$}}

\usepackage[makeroom]{cancel}

\newtheorem{remark}{Remark}

\title{Who is Afraid of Big Bad Minima? \\ Analysis of Gradient-Flow in a Spiked Matrix-Tensor Model}

\author[1]{Stefano Sarao Mannelli}
\author[2]{Giulio Biroli}
\author[3]{Chiara Cammarota}
\author[2]{Florent Krzakala}
\author[1]{Lenka Zdeborov\'a}
\affil[1]{Institut de physique th\'eorique, Universit\'e Paris Saclay, CNRS, CEA, 91191 Gif-sur-Yvette, France}
\affil[2]{Laboratoire de Physique de l'Ecole normale sup\'erieure ENS,  Universit\'e PSL, CNRS, Sorbonne Universit\'e, Universit\'e Paris-Diderot, Sorbonne Paris Cit\'e Paris, France}
\affil[3]{Department of Mathematics, King's College London, Strand London WC2R 2LS, UK}

\date{}

\begin{document}

\maketitle

\begin{abstract}
	Gradient-based algorithms are effective for many machine learning tasks, but despite ample recent effort and some progress, it often remains unclear why they work in practice in optimising high-dimensional non-convex functions and why they find good minima instead of being trapped in spurious ones.
	Here we present a quantitative theory explaining this behaviour in a spiked matrix-tensor model.
	Our framework is based on the Kac-Rice analysis of stationary points and a closed-form analysis of  gradient-flow originating from statistical physics. We show that there is a well defined region of parameters where the gradient-flow algorithm finds a good global minimum despite the presence of exponentially many spurious local minima.
	We show that this is achieved by surfing on saddles that have strong negative direction towards the global minima, a phenomenon that is connected to a BBP-type threshold in the Hessian
	describing the critical points of the landscapes.
\end{abstract}


\section{Introduction}

A common theme in machine learning and optimisation is to understand the behaviour of gradient descent methods for non-convex problems with many minima. Despite the non-convexity, such methods often successfully optimise models such as neural networks, matrix completion and tensor factorisation. This has motivated a recent spur in research attempting to characterise the properties of the loss landscape that may shed some light on the reason of  such success. Without the aim of being exhaustive these include \cite{kawaguchi2016deep,soudry2016no,ge2016matrix,freeman2016topology,bhojanapalli2016global,park2016non,du2017gradient,ge2017optimization,ge2017no,lu2017depth,ling2018landscape}.

Over the last few years, a popular line of research has shown, for a variety of systems, that spurious local minima are not present in certain regimes of parameters. When the signal-to-noise ratio is large enough, the success of gradient descent  can thus be understood by a trivialisation transition in the loss landscape: either there is only a single minima, or all minima become "good", and no  spurious minima can trap the dynamics. This is what happens, for instance, in the limit of small noise and abundance of data for matrix completion and tensor factorization \cite{ge2016matrix,ge2017optimization}, or for some very large neural networks \cite{kawaguchi2016deep,soudry2016no}.
However, it is often observed in practice that these guarantees fall short of explaining the success of gradient descent, that is empirically observed to find good minima very far from the regime under mathematical control. In fact, gradient-descent-based algorithms may be able to perform well even when spurious local minima are present because the basins of attraction of the spurious minima may be small and the dynamics might be able to avoid
them. Understanding this behaviour requires, however, a very detailed characterisation of the dynamics and of the landscape, a feat which is not yet possible in full generality.

A fruitful direction is the study of
Gaussian functions on the $N$-dimensional sphere, known as $p$-spin spherical spin glass
models in the physics literature, and as isotropic models in the Gaussian process literature~\cite{gross1984simplest,fyodorov2004complexity,auffinger2013random,sagun2014explorations,arous2017landscape}. In statistics and machine learning, these models have appeared following the studies of spiked matrix and tensor models \cite{johnstone2009consistency,deshpande2014information,richard2014statistical}. In particular, a very recent work \cite{sarao2019passed} showed explicitly that for a spiked matrix-tensor model the gradient-flow algorithm indeed reaches global minimum even when spurious local minima are present and the authors estimated numerically the corresponding regions of parameters.
In this work we consider this very same model and explain the mechanism by which the spurious local minima are avoided, and develop a quantitative theoretical framework that we believe has a strong potential to be generic and extendable to a much broader range of models in high-dimensional inference and neural networks.

%
{\bf The Spiked Matrix-Tensor Model}. The spiked matrix-tensor model has been recently proposed to be a prototypical model for non-convex high-dimensional optimisation where several non-trivial regimes of cost-landscapes can be displayed quantitatively by tuning the parameters \cite{sarao2018marvels,sarao2019passed}. 
In this model, one aims at reconstructing a hidden vector (i.e. the spike) $\pmb\sigma^*$ from the observation of a noisy version of {\it both} the rank-one matrix and rank-one tensor created from the spike. Using the following notation: bold lowercase symbols represent vectors, bold uppercase symbols  represent matrices or tensors, and $\langle \cdot,\cdot \rangle$  represent the scalar product, the model is defined as follows: given a signal (or spike), $\pmb\sigma^*$, uniformly sampled on the $N$-dimensional hyper-sphere of radius $1$, it is given a tensor $\pmb T$ and a matrix $\pmb Y$ such that
\begin{align}\label{eq:observationT}
	 & T_{i_1\dots i_p} = \eta_{i_1\dots i_p} + \sqrt{N(p-1)!}\, \sigma^*_{i_1}\dots \sigma^*_{i_p}\,, \\
	\label{eq:observationY}
	 & Y_{ij} = \eta_{ij} + \sqrt N \sigma^*_i \sigma^*_j\,,
\end{align}
where $\eta_{i_1\dots i_p}$ and $\eta_{ij}$ are Gaussian random variables of variance $\Delta_p$ and $\Delta_2$ respectively. 
Neglecting constant terms, the
maximum likelihood estimation of the ground truth, $\pmb\sigma^*$, corresponds to minimization of the following loss function:
\begin{equation} \label{eq:loss}
	\loss(\pmb\sigma | \pmb T, \pmb Y) = -\frac{\sqrt{(p-1)!}}{\Delta_p\sqrt{N}}\sum_{i_1<\dots<i_p} T_{i_1\dots i_p} \sigma_{i_1}\dots\sigma_{i_p} - \frac1{\Delta_2\sqrt{N}}\sum_{i<j} Y_{ij} \sigma_i \sigma_j=
\end{equation}
$$
	-\frac{\sqrt{(p-1)!}}{\Delta_p\sqrt{N}}\sum_{i_1<\dots<i_p} \eta_{i_1\dots i_p} \sigma_{i_1}\dots\sigma_{i_p} - \frac1{\Delta_2\sqrt{N}}\sum_{i<j} \eta_{ij} \sigma_i \sigma_j
	-\frac{\langle\pmb\sigma, \pmb \sigma^*\rangle^p}{p\Delta_p} 
	-\frac{\langle\pmb\sigma, \pmb \sigma^*\rangle^2}{2\Delta_2}\,.
$$
The first two contributions of the last equation will be denoted $\epsilon_p$ and $\epsilon_2$ in the following.
While the matricial observations correspond to a quadratic term and thus to a simple loss-landscape, the additional order-$p$ tensor contributes towards a rough non-convex loss landscape.
As $\Delta_p\!\rightarrow\!\infty$ the information $T_{i_1\dots i_p}$  becomes irrelevant and the landscape  becomes trivial,  while in the opposite limit $\Delta_2\!\rightarrow\!\infty$, the landscape  becomes extremely rough and complex as analyzed recently in \cite{arous2017landscape,ros2018complex}.

We shall consider the behaviour of the gradient-flow (GF) algorithm aiming to minimise the loss:
\begin{equation}\label{eq:gradient_flow}
	\dot\sigma_i(t) = -\mu(t)\sigma_i(t) - \frac{\partial \loss(\pmb\sigma(t) | \pmb T, \pmb Y)}{\partial \sigma_i(t)}\,,
\end{equation}
where $\mu(t)$ enforces that  $\pmb\sigma(t)$ belongs to the hyper-sphere of radius $N$ and will be referred to as the {\it spherical constraint}.
%
The algorithm is initialised in a random point drawn uniformly on the hyper-sphere, thus initially having no correlation with the ground-truth signal.
We view the gradient-flow as a prototype of gradient-descent-based algorithms that are the work-horse of current machine learning.

\begin{figure}[t!]
	\centering
	\includegraphics[width=0.85\linewidth]{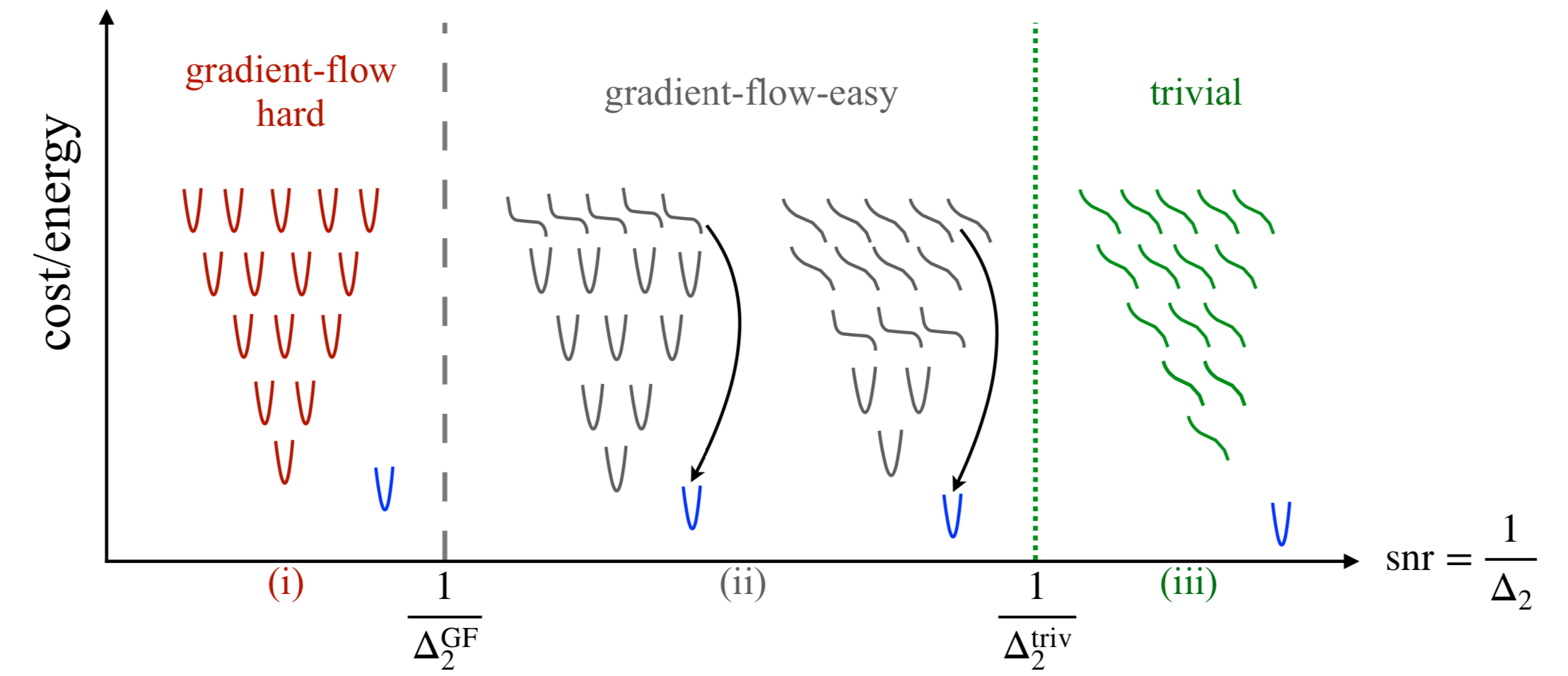}  
	\caption{Cartoon illustrating the mechanism by which gradient-flow avoids spurious local minima in the spiked matrix-tensor model.	As the signal-to-noise ($\rm snr$) ratio $1/\Delta_2$ is increased, the spurious local minima that attract the randomly initialised GF algorithm develop a single negative direction towards the global minimum before the others, in particular the lower-cost spurious local minima, do. This has drastic consequences on the GF algorithm. In region (i), for  ${\rm snr}<1/\Delta_2^{\rm GF}$, the algorithm goes down the landscape, eventually reaches the high-energy {\it threshold} minima and remains stuck. In region (ii), however, these threshold minima are turned into saddles with a strong negative direction towards the signal. The algorithm is initially reaching  these minima-turned-saddles, surfing on the negative slope, it then turns towards the "good" minima correlated with the signal, avoiding the exponentially many spurious minima at lower energies.
	The main technical contribution of this paper is a quantitative description of this scenario, including a simple formula for the corresponding threshold $\Delta_2^{\rm GF}$, eq.~(\ref{eq:GD_threshold_kac-rice}). As the {\rm snr} is further increased, the negative direction appears in lower and lower minima until the trivialization transition in region (iii): for ${\rm snr}>1/\Delta_2^{\rm triv}$, all the spurious minima have been turned into saddles.
	\label{fig:cartoon}
	}
\end{figure}
{\bf Main Contributions}.
The first main result of this paper is the expression for the threshold
below which the gradient-flow algorithm finds a configuration correlated the hidden spike. This threshold is established in the asymptotic limit of large $N$, fixed $p$ and $\Delta_p$, and reads:
\begin{equation}\label{eq:GD_threshold_kac-rice}
	\Delta^{\rm GF}_2 (\Delta_p,p) \equiv \frac{-\Delta_p+\sqrt{\Delta_p^2+4(p-1)\Delta_p}}{2(p-1)}\,.
\end{equation}
We find that  (i) for $\Delta_2<\Delta^{\rm GF}_2$ the gradient flow reaches in finite time the global minimum, well correlated with the signal, while (ii) for  $\Delta_2>\Delta^{\rm GF}_2$ the algorithm remains uncorrelated with the signal for all times that do not grow as $N$ grows. We contrast it with the threshold $ \Delta_2^{\rm triv} < \Delta^{\rm GF}_2 $, established in \cite{sarao2019passed}, below which the energy landscape does not present any spurious local minima. Note that
$\Delta^{\rm GF}_2 $ is less than $\Delta_2^{\rm AMP}=1$ \cite{sarao2018marvels}, below which the best known algorithm, specifically the approximate message passing, works.

The second main result of this paper, 
is the insight we obtain on the behaviour of the gradient-flow in the loss landscape, that is summarised in Fig.~\ref{fig:cartoon}. The key point is to consider the fate of the spurious local minima that attract the GF algorithm when the signal to noise ratio ${\rm snr}=1/\Delta_2$ is increased. As the ${\rm snr}$ increases, these minima turn into saddles with a single negative direction towards the signal (a phenomenon that we analyze in the next section, and that turns out to be  linked to the BBP transition \cite{baik2005phase} in random matrix theory), all that well before all the other spurious local minima disappear. We present two ways to quantify this insight:

(a) We use the Kac-Rice formula for the number of stationary points, as derived for the present model in \cite{sarao2019passed}. In \cite{sarao2019passed} this formula is used to quantify the region with no spurious local minima. Here we focus on a BBP-type of phase transition that is crucial in the derivation of this formula and deduce the GF threshold (\ref{eq:GD_threshold_kac-rice}) from it.

(b) We use the CHSCK equations \cite{CHS93,cugliandolo1993analytical} for closed-form description of the behaviour of the gradient-flow, as derived and numerically solved in \cite{sarao2018marvels,sarao2019passed}. Building on dynamical theory of mean-field spin glasses we determine precisely when and how the algorithm escapes the manifold of zero overlap with the signal, leading again to the threshold (\ref{eq:GD_threshold_kac-rice}).

Both these arguments are derived using reasoning common in theoretical physics. From a mathematically rigorous point of view the threshold (\ref{eq:GD_threshold_kac-rice}) remains a  conjecture and its rigorous proof is an interesting challenge for future work. We note that both the Kac-Rice approach \cite{arous2017landscape} and the CHSCK equations \cite{arous2006cugliandolo} have been made rigorous in closely related problems. We believe that the reported results are not limited to the present model and we will investigate analytically and numerically other models and real-data-based learning in order to validate this theory and to understand its limitations.




\begin{figure}[ht!]
	\centering
	\includegraphics[width=0.9\linewidth]{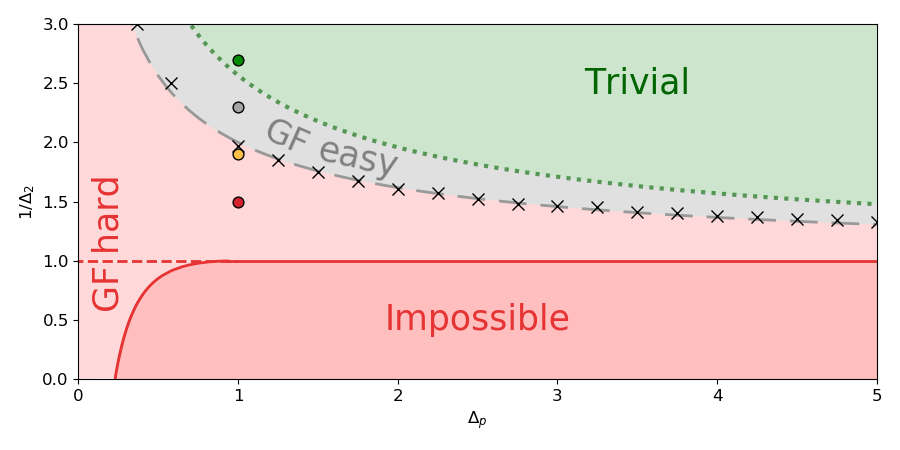}
	\caption{The phase diagram shows the different regions for gradient-flow behaviour, for the spiked matrix-tensor model with $p=3$. In the region shaded in red (light and dark), GF does not correlate with the signal, while it does in the grey and green regions. In the dark-red region obtaining correlation with the signal is \textit{impossible} information-theoretically \cite{sarao2018marvels}.
		The possible region is divided by a red dashed line, below that line even best known algorithms are unable to obtain correlation with the signal \cite{sarao2018marvels}.
		The green region is characterised by a \textit{trivial} landscape, i.e. all the spurious minima disappear \cite{sarao2019passed}. The grey region is where gradient-flow succeeds to converge despite the presence of spurious minima. We marked with black crosses points predicting the gradient-flow threshold obtained numerically in \cite{sarao2019passed}, they perfectly agree with our theoretical prediction of the threshold (\ref{eq:GD_threshold_kac-rice}), marked by the grey dashed line. The circles in colours are points that we will use to illustrate the different features of these regions.
	}
	\label{fig:phase_diagram}
\end{figure}

\section{Probing the Landscape by the Kac-Rice Method}\label{sec:Kac-Rice}

The statistical properties of the landscape associated to the loss function (\ref{eq:loss}) can be studied by the Kac-Rice method, which traces back to the statistical physics literature, see \cite{CC05} for an overview, and was developed mathematically
in \cite{fyodorov2004complexity,auffinger2013random} and recently extended in \cite{ros2018complex}.

The quantities of interest are the number of critical points at a given energy, ${\mathcal N}(\epsilon_p,\epsilon_2)$, and the Hessian matrix evaluated at those critical points.
We analyse the logarithm of ${\mathcal N}(\epsilon_p,\epsilon_2)$, called the {\it complexity}. Since the complexity is a random quantity we compute its upper bound $\Sigma_a(\epsilon_p,\epsilon_2)=\ln \mathbb{E}[{\mathcal N}(\epsilon_p,\epsilon_2)]$, along the lines of \cite{arous2017landscape,sarao2019passed}. We have also computed its typical value $\Sigma_q(\epsilon_p,\epsilon_2)=\mathbb{E}[\ln {\mathcal N}(\epsilon_p,\epsilon_2)]$ along the lines of \cite{ros2018complex}, i.e. non-rigorously using the replica symmetry assumption (see SM Sec.~\ref{sec:kac-rice}). In what follows we focus on complexity of stationary points with no correlation with the signal, in which case analytical and numerical arguments (see SM Sec.~\ref{sec:kac-rice_joint_density}) indicate
that $\Sigma_a(e_p,e_2)$ and $\Sigma_q(e_p,e_2)$ are either very close numerically or possibly equal. Thus, in the following, we will simply refer
to the complexity $\Sigma(\epsilon_p,\epsilon_2)$ without further specification.

In the Kac-Rice analysis the statistics of the Hessian, $\mathcal{H}$, of critical points plays a key role. It was shown in \cite{sarao2019passed}, and the argumentation is reproduced in the SM Sec.~\ref{sec:kac-rice_derivation}, that $\mathcal{H}$ has a simple form for the loss (\ref{eq:loss}). It is a $(N-1)\times (N-1)$ matrix formed by the sum of three contributions: a random matrix $\mathbb{W}_{N-1}$ belonging to the Gaussian orthogonal ensemble (GOE), a matrix proportional to the identity, and a rank one projector in the direction of the signal. The expression of $\mathcal{H}$ for critical points with null overlap $m$ with the signal and with energies $\epsilon_p$ and $\epsilon_2$ reads:
\begin{equation}\label{eq:Hessian}
	\mathcal{H} = \sqrt{Q''(1)} \left[\mathbb{W}_{N-1} + t\,\mathbb{I}_{N-1} - \theta\,\pmb{e}_1\pmb{e}_1^T\right]
\end{equation}
with $Q(x) = \frac{x^p}{p\Delta_p} + \frac{x^2}{2\Delta_2}$, $t = -\left(p\epsilon_p+2\epsilon_2\right)/\sqrt{Q''(1)}$, and $\theta = Q''(0)\,/\sqrt{Q''(1)}$. The normalisation of $\mathbb{W}_{N-1}$ is chosen such that $\mathrm{Tr}\mathbb{E}[\mathbb{W}_{N-1}^2]=1$.

{\bf The Fate of the Spurious:} The initial condition for the gradient-flow algorithm is a random configuration $\bs$ uniformly drawn on the hyper-sphere. Such an initial condition 
clearly belongs to the large manifold of configurations uncorrelated with the ground-truth signal.
We aim to investigate how does the gradient flow manage to escape from this initial manifold.
For this purpose we focus on the properties of the landscape in the subspace where the overlap with the signal is zero, $m=0$.

\begin{figure}[t!]
	\centering
	\includegraphics[scale=.43]{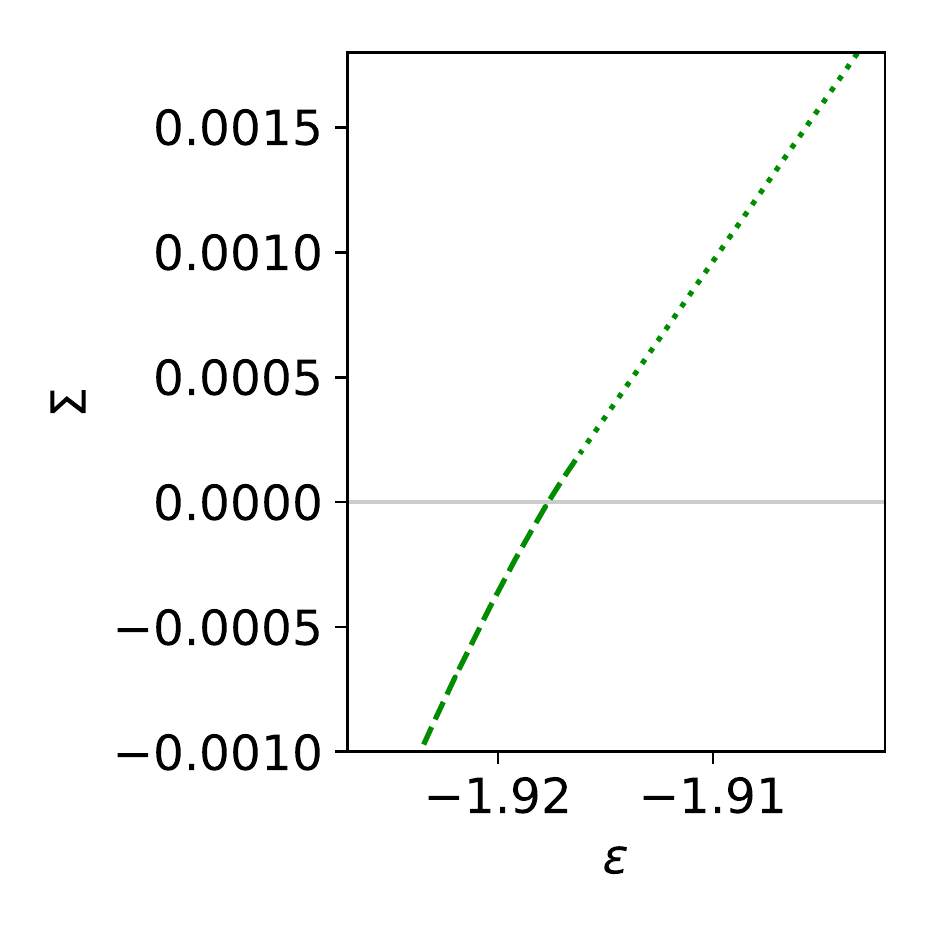}
	\includegraphics[scale=.43]{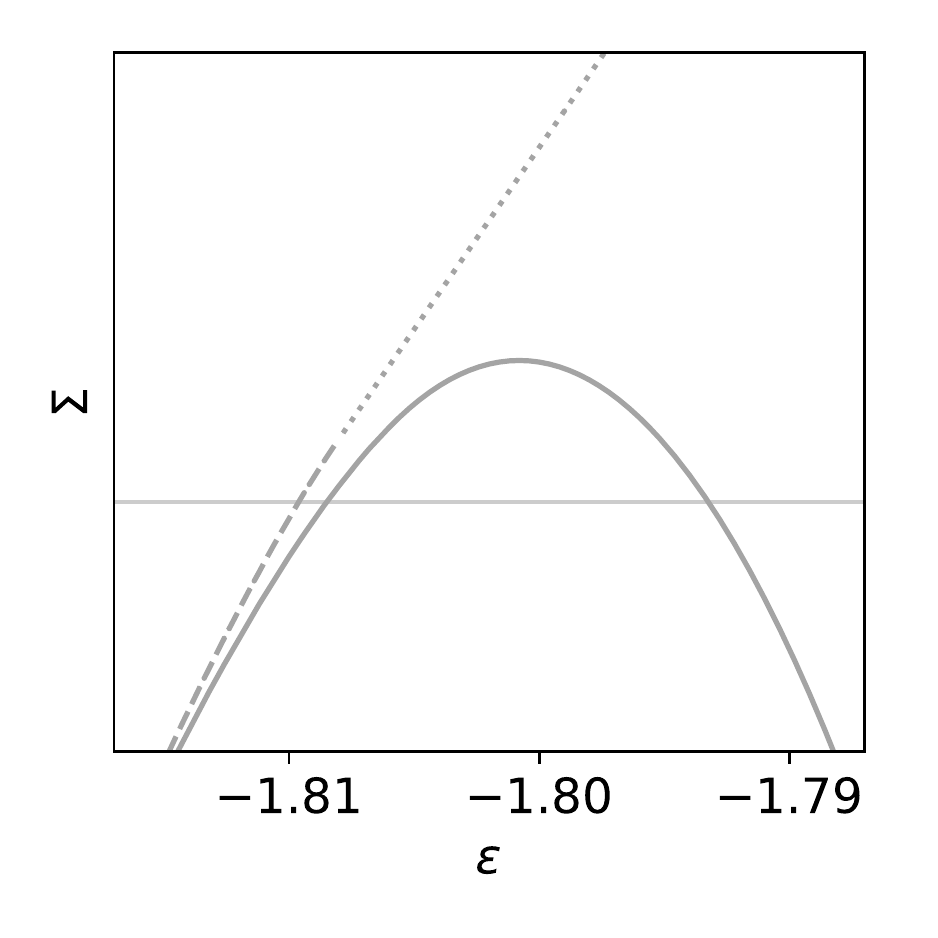}
	\includegraphics[scale=.43]{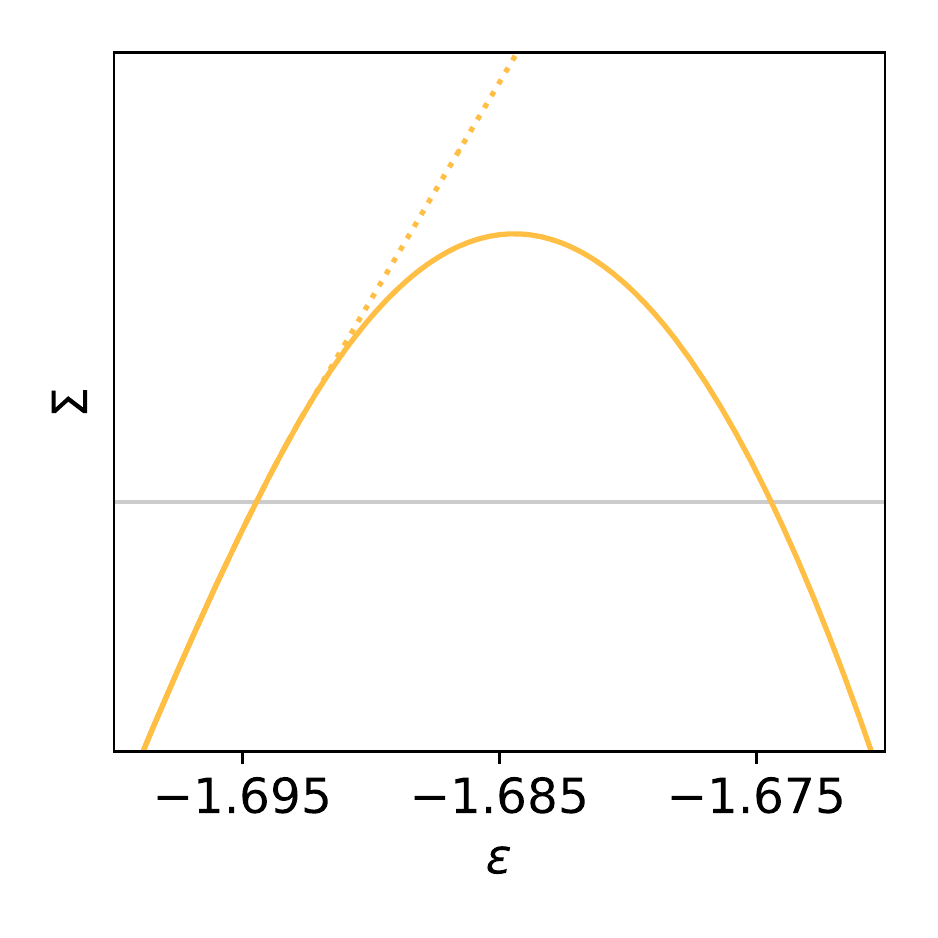}
	\caption{Complexity curves for the number of critical points for an overlap value $m=0$ at fixed $\Delta_p = 1.0$ for (from left to right) $1/\Delta_2=2.7, 2.3, 1.9$. The lines are dotted when the complexity is dominate by critical points having an extensive number of eigenvalues are negative, dashed when only one eigenvalue is negative, full when the points have only positive (or null) eigenvalues i.e. they are minima. The complexity of the minima is drawn in full lines with the same colours, and it merges with the complexity of stationary points when it becomes dominant.}
	\label{fig:complexity_energy_zoom}
\end{figure}

In Fig.~\ref{fig:complexity_energy_zoom}, we plot the complexity at $m=0$ as a function of the energy $\epsilon$
\begin{equation*}\label{sigma}
	\Sigma(\epsilon)=\sup_{\substack{\epsilon_p,\epsilon_2\\\text{s.t. }\epsilon_p+\epsilon_2=\epsilon}}\Sigma(\epsilon_p,\epsilon_2)\Big|_{m=0}
\end{equation*}
for the points $1/\Delta_2=1.9, 2.3, 2.7$ and $\Delta_p = 1.0$  ($p=3$), which are marked with circles of the corresponding colour in Fig.~\ref{fig:phase_diagram}.
We use discontinuous lines for the complexity of critical points that have at least one negative direction, and full lines for the complexity of local minima.
A finding of \cite{sarao2019passed}, that holds for any value of $\Delta_p$, is that for small $1/\Delta_2$ the majority of critical points with zero overlap with the signal at low enough energies are spurious minima; they disappear increasing $1/\Delta_2$ above a $\Delta_p$-dependent value $1/\Delta^{\rm triv}_2$ corresponding to the green region of Fig.~\ref{fig:phase_diagram}. In this part of the phase diagram, there are no spurious minima and the global minimum is correlated with the signal; this is an "easy" landscape for gradient flow which is therefore expected to succeed there.
The main open question concerns the behavior for smaller values of $1/\Delta_2$: When does the existence of spurious minima, appearing in panel (b) and (c) of Fig.~\ref{fig:complexity_energy_zoom}, start to be  harmful to gradient flow?

In order to answer this question, we investigate more closely the nature of the spurious minima at different energies. We focus in particular on their Hessian, which 
plays a crucial role in order to understand which spurious minima have the largest basin of attraction and, hence, can trap the algorithm. For low signal-to-noise ratio, large $\Delta_p$ and large $\Delta_2$, the spectrum of (\ref{eq:Hessian})  is a shifted Wigner semicircle with support $[\sqrt{Q''(1)}(-2+t),\sqrt{Q''(1)}(2+t)]$. The effect of the projector (third contribution to the RHS of (\ref{eq:Hessian})) on the support of the spectrum is negligible as long as $\theta\le 1$, as follows from the work on low-rank perturbations of random GOE matrices \cite{baik2005phase}. Moreover, the most numerous critical points at fixed energy $\epsilon $ are characterized by a $t(\epsilon)$ that is a monotonously decreasing function of $\epsilon$, see Fig.~\ref{fig:t_energy} in the SM. Thus, moving towards higher energies, the spectrum of the Hessian shifts to the left, which indicates smaller curvature and wider minima. 
The transition between minima and saddles takes place at the {\it threshold energy} at which $t(\epsilon_{\rm th})=2$, i.e. where the left edge of the Wigner semi-circle law touches zero, the numerical value is obtained in the Appendix Sec.~\ref{sec:kac-rice_complexities}. Above this energy, critical points have an extensive number of downward directions, as found also in spin-glass models \cite{auffinger2013random,CC05}.
Putting the above findings together, minima at $\epsilon=\epsilon_{\rm th}$ are the most numerous and {\it the marginally stable ones}. Therefore, they are the natural candidates for having the largest basin of attraction and the highest influence on the randomly initialised algorithm. This reasonable guess is at the basis of the theory of glassy dynamics in physics \cite{cugliandolo1993analytical}. We take it as a working hypothesis for now, and we confirm it analytically and numerically in what follows.

Going back to the algorithm, when the signal-to-noise ratio is small we therefore expect that the configuration $\pmb\sigma(t)$ slowly approaches at long times the ubiquitous "threshold minima" characterised by energy $\epsilon_{\rm th}$ and zero overlap with the signal.
The last missing piece is unveiling what makes those minima unstable for large $\rm snr$. We show below that it is a transition, called BBP (Baik-Ben Arous-Péché) \cite{baik2005phase}, which takes place in the spectrum of the Hessian.
Increasing $1/\Delta_2$ at fixed $\Delta_p$, as in Fig.~\ref{fig:complexity_energy_zoom}, leads to a larger $\theta$. When $\theta$ becomes larger than one, an eigenvalue, equal to $\sqrt{Q''(1)}\left(-\theta- \theta^{-1}+t\right)$ pops out on the left of the Wigner semi-circle, and its corresponding eigenvector develops a finite overlap with the signal \cite{baik2005phase}. This 
implies
the development of an unstable direction for the threshold minima trapping the dynamics, as they were already at the verge of instability. Indeed, as soon as the isolated eigenvalue pops out, an unstable downward direction towards the signal emerges in the landscape around them, at which  point  the algorithmic threshold for gradient flow takes place. Interestingly, 
many other spurious minima at lower energy also undergo the BBP transition, but they remain stable for longer as the isolated eigenvalue is positive when it pops out from the semi-circle.
In conclusion, our analysis of the landscape suggests a dynamical transition for signal estimation by gradient flow given by
\begin{equation}\label{BBP}
	\theta=Q''(0)/\sqrt{Q''(1)} = 1
\end{equation}
which leads to a very simple expression for the transition line $\Delta^{\rm GF}_2$, Eq.~\eqref{eq:GD_threshold_kac-rice}.
This theoretical prediction is shown in Fig.~\ref{fig:phase_diagram} as a dashed grey line: The agreement with the numerical estimation from \cite{sarao2019passed} (black crosses) is perfect.

Our analysis unveils that the key property of the loss-landscape determining the performance of the gradient-flow algorithm, is the (in)stability in the direction of the signal of the minima with largest basin of attraction. These are the most numerous and the highest in energy, a condition that likely holds for many high-dimensional estimation problems.

The other spurious minima, which are potentially more trapping than the threshold ones and still stable at the algorithmic transition just derived, are actually completely innocuous since a random initial condition does not lie in their basin of attraction with probability one in the large $N$ limit.
This benign role of very bad spurious minima might appear surprising; it is due to the high-dimensionality of the non-convex loss function. Indeed it does not happen in finite dimensional cases, in which a random initial condition has instead a finite probability to fall into bad minima if those are present.

\section{Probing the Gradient-Flow Dynamics}\label{sec:CHSCK}

\subsection{Closed-Form Dynamical Equations}

In the large $N$ limit gradient-flow dynamics for the spiked matrix-tensor model can be analysed using techniques originally developed in statistical physics studies of spin-glasses \cite{crisanti1992sphericalp,CK94SK,CK95} and later put on a rigorous basis in \cite{arous2006cugliandolo}. Three observables play a key role in this theory:

(i) The overlap (or correlation) of the estimator at two different times: $C(t,t') = \langle \pmb\sigma(t), \pmb\sigma(t') \rangle$.

(ii) The change (or response) of the estimator at time $t$ due to an infinitesimal perturbation in the loss at time $t'$, i.e. $\loss\rightarrow \loss+\langle\pmb\sigma(t'), \pmb h(t')\rangle$ in Eq.~\eqref{eq:gradient_flow}: $R(t,t') = \sum_{i=1}^N \frac{\delta \sigma_i(t)}{\delta h_i(t')}\Big|_{h_i = 0}$.

(iii) The average overlap of the estimator with the ground truth $m(t) = \langle \pmb\sigma^*,\pmb\sigma(t)\rangle$.

For $N\rightarrow \infty$ the above quantities converge to a non-fluctuating
limit, i.e. they concentrate with respect to the randomness in the initial condition and in the generative process, and satisfy closed equations.
Following works of Crisanti-Horner-Sommers-Cugliandolo-Kurchan (CHSCK) \cite{CK94SK,CK95} and their recent extension to the spiked matrix-tensor model \cite{sarao2018marvels,sarao2019passed} the above quantities satisfy:
\begin{align}
	\begin{split}
		&\frac{\partial}{\partial t} C(t,t') = - \mu(t)\,C(t,t')+
		Q'(m(t)) m(t') + \int_0^t
		R(t,t'')Q''(C(t,t''))C(t',t'') dt''
		\\
		&\quad+ \int_0^{t'} R(t',t'')Q'(C(t,t'')) dt''\,,
	\end{split}\label{eq:CHSCK_C}
	\\
	\begin{split}
		&\frac{\partial}{\partial t} R(t,t')= - \mu(t)\,R(t,t') +\int_{t'}^t R(t,t'')Q''(C(t,t''))R(t'',t') dt''\,,
	\end{split}\label{eq:CHSCK_R}
	\\
	\begin{split}
		&\frac{d}{d t} m(t) = -\mu(t)\,m(t)+Q'(m(t)) + \int_{0}^t R(t,t'')m(t'') Q''(C(t,t'')) dt''\,,
	\end{split}\label{eq:CHSCK_Cbar}
	\\
	\begin{split}
		& \mu(t) = Q'(m(t))m(t) + \int_{0}^t R(t,t'')\left[Q'(C(t,t'')) + Q''(C(t,t''))\,C(t,t'')\right] dt''\,,
	\end{split}\label{eq:CHSCK_mu}
\end{align}
with initial conditions $C(t,t)=1\;\forall t$ and $R(t,t')=0$  for all $t<t'$ and
$\lim_{t'\rightarrow t^-}R(t,t')=1\;\forall t$.
The additional function $\mu(t)$, and its associated equation, are due to the spherical constraint; $\mu(t)$ plays the role of a Lagrange multiplier and guarantees that the solution of the previous equations is such that $C(t,t)=1$.
The derivation of these equations can be found in \cite{sarao2018marvels} and in the SM Sec.~\ref{sec:CHSCK_appendix}.
It is obtained using heuristic theoretical physics approach and
can be very plausibly made fully rigorous generalising the work of
\cite{arous2006cugliandolo,zamfir2008limiting}.

This set of equations can be solved numerically as described in \cite{sarao2018marvels}. The numerical estimation of the algorithmic threshold of gradient-flow, reproduced in Fig.~\ref{fig:phase_diagram}, was obtained in \cite{sarao2019passed}.  We have also directly simulated the gradient flow Eq.~\eqref{eq:gradient_flow} and compare the result to the one obtained
from solving Eqs.~(\ref{eq:CHSCK_C}-\ref{eq:CHSCK_mu}).
As shown in the SM Sec.~\ref{sec:numerical_study}, for $N=65535$, we find a very good agreement even for this large yet finite size.



\begin{figure}[t!]
	\centering
	\includegraphics[scale=0.5]{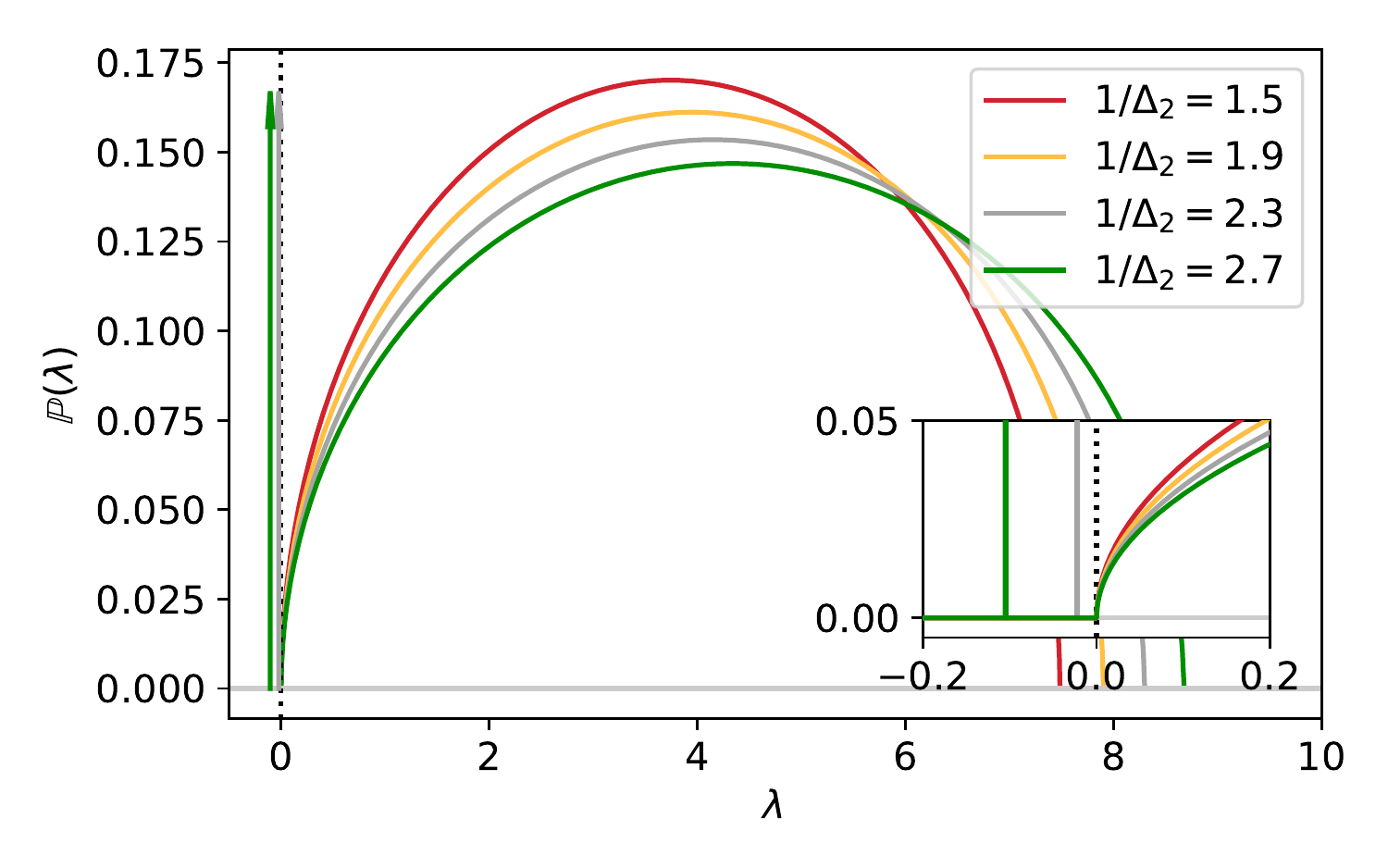}
	\includegraphics[scale=0.5]{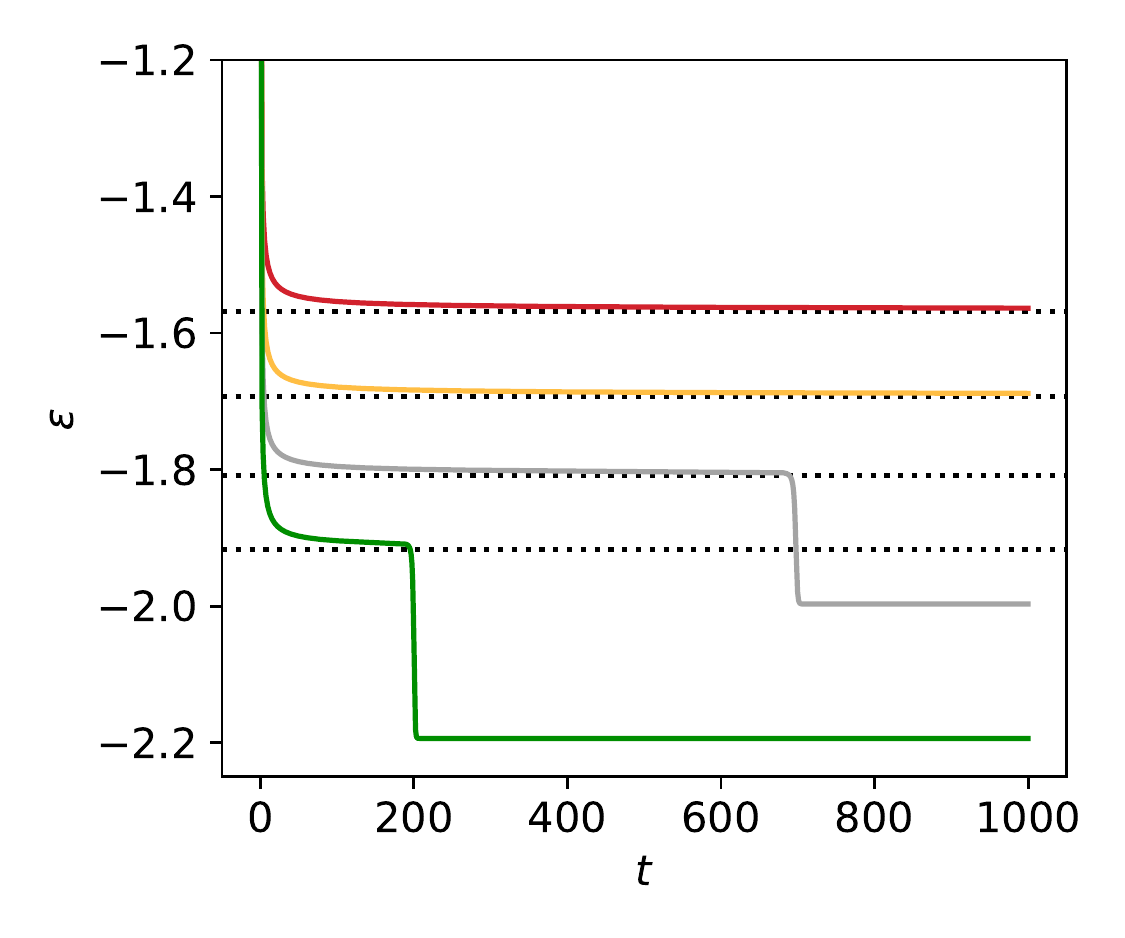}
	\caption{Right panel: energy as a function of time for the set of parameters indicated by small circles in Fig.~\ref{fig:phase_diagram}. The horizontal dotted lines correspond to value of the threshold energy $\epsilon_{\rm th}$, as derived both from the Kac-Rice approach in Appendix Sec.~\ref{sec:kac-rice_complexities} and from the large time behaviour of the dynamics in Appendix Sec.~\ref{sec:parameters}. Left panel: Eigenvalue distribution of the Hessian of the threshold states for the same set of parameters. When $1/\Delta_2$ becomes smaller than $2$ an isolated eigenvalue appears; it has been highlighted using vertical arrows. Concomitantly, the energy as a function of time first approaches the plateau and eventually departs from it and reaches the energy of the global minimum.
	}
	\label{fig:dynamics_complexity}
\end{figure}

{\bf Surfing on saddles:} Armed with the dynamical equations, we now 
confirm the prediction of the threshold (\ref{eq:GD_threshold_kac-rice}) based on the Kac-Rice-type of landscape analysis. In the SM we check that 
the minima trapping the dynamics are indeed the marginally stable ones ($t=2$), see Figs.~\ref{fig:t_ag} and \ref{fig:hess_ag} in the SM, and we show the energy can be expressed in terms of $C$, $R$ and $m$.
In the right panel of Fig.~\ref{fig:dynamics_complexity} we then plot the energy as a function
of time obtained from the numerical solution of Eqs.~(\ref{eq:CHSCK_C}-\ref{eq:CHSCK_mu}) for $1/\Delta_2=1.5, 1.9, 2.3, 2.7$ and $\Delta_p=1$ (same points and colour code of Figs.~\ref{fig:phase_diagram} and \ref{fig:complexity_energy_zoom}). 
For the two smaller values of $1/\Delta_2$ the energy converges to a plateau value at $\epsilon_{\rm th}$ (dotted line), whereas for $1/\Delta_2=2.3, 2.7$ the energy plateaus close to $\epsilon_{\rm th}$  but then eventually drifts away and reaches a lower value, corresponding to the global minimum correlated with the signal.
This behaviour can be understood
in terms of the spectral properties of the Hessian (\ref{eq:Hessian}) of the minima 
trapping the dynamics. 
In the left panel of Fig.~\ref{fig:dynamics_complexity} we plot the corresponding density of
eigenvalues of $\mathcal H$ for the same values of $1/\Delta_2$ and $\Delta_p$ used in the right panel. This is an illustration of the dynamical phenomenon explained in the previous section: when the signal-to-noise ratio is large enough threshold minima become unstable because a negative eigenvalue, associated to a downward direction toward the signal, emerges. In this case $\pmb \sigma (t)$ first seems to converge to the threshold minima and then, at long times, drifts away along the unstable direction. The larger is the signal-to-noise ratio the more unstable is the downward direction and, hence, the shortest is the intermediate trapping time.



\subsection{Gradient-flow Threshold from Dynamical Theory}\label{sec:CHSCK_threshold}

We now show that the very same prediction (\ref{eq:GD_threshold_kac-rice}) for the algorithmic threshold of gradient-flow can be directly obtained analysing the dynamical equations (\ref{eq:CHSCK_C}-\ref{eq:CHSCK_mu}), without directly using results from the Kac-Rice analysis,
thus establishing a firm and novel connection between the behaviour of the gradient-flow algorithm and Kac-Rice landscape approaches.

For small signal-to-noise ratios, when $m$ remains zero at all times, the dynamical equations (\ref{eq:CHSCK_C}-\ref{eq:CHSCK_mu}) are identicall to the well-known one in   spin glasses theory, for reviews see \cite{bouchaud1998out,Cu03}. These equations have been studied extensively for decades in statistical physics and a range of results about their behaviour has been established.
Here we describe the results which are important for our analysis and devote the SM Sec.~\ref{sec:CHSCK_CK_analysis} to a more extended presentation. 
It was shown analytically in \cite{CK94SK} that the behaviour of the dynamics at large times is captured by an asymptotic solution of Eqs.~(\ref{eq:CHSCK_C}-\ref{eq:CHSCK_mu}) that verifies several remarkable properties.
The ones of interest to us are that for $t$ and $t'$ large:

(i) $C(t,t')=1$ when $t-t'$ finite; $C(t,t')$ becomes less than one when $t-t'$ diverges with $t$ and $t'$.

(ii) $R(t,t')=R_{\rm TTI}(t-t')+R_{\rm ag}(t,t')$, where TTI stands for time-translational-invariance, ag stand for aging. Here $R_{\rm TTI}(t-t')$ goes to zero on a finite time-scale, whereas $R_{\rm ag}(t,t')$ varies
on timescales diverging with $t$ and $t'$. Moreover, $R_{\rm ag}(t,t')$
verifies the
so called "weak-long term memory" property:
for any finite $t_0$, $\int^t_{t-t_0}R_{\rm ag}(t,t'')dt''$
is arbitrarily small. We refer to this function form for $R(t,t')$ as the {\it aging ansatz}, adopting the physics terminology.

These properties are confirmed to hold by our numerical solution, see for instance Fig.~\ref{fig:correlation_fdt} in the SM. The interpretation of these dynamical properties is that at long times $\pmb\sigma(t)$ decreases in the energy landscape and approaches the marginally stable minima.
Concomitantly, dynamics slows down and takes place along the almost flat directions associated to the vanishing eigenvalues of the Hessian.

We remind that in the previous paragraphs we assume null correlation with the signal, $m=0$.
In order to find the algorithmic threshold beyond which the gradient-flow develops a positive correlation, we study the instability of the aging solution as a function of the signal-to-noise ratio.
Our strategy is to start with an arbitrarily small overlap, $m(0)=\delta$, and determine
whether it grows at long times thus indicating an instability towards the signal. Since the initial condition for the overlap is uncorrelated with the signal, then, for sufficiently small $\delta$, $C$ and $R$ reach their asymptotic form before $m$
becomes of order one.
We can thus plug the asymptotic aging ansatz for $R$ in the dynamical equation for $m$:
\begin{equation}\label{eq:mag}
	\begin{split}
		\frac{d}{dt}m(t) &= -\mu(t)m(t) +Q'(m(t)) + \int_{0}^t R_{\rm TTI}(t-t'')Q''(1)m(t'') dt'' +
		\\
		&+\int_0^{t} R_{\rm ag}(t,t'')Q''(C_{\rm ag}(t,t''))m(t'') dt'' \
	\end{split}
\end{equation}
In the linear approximation the solution has the form $m(t)=\delta \exp(\Lambda t)$ and we assume $\Lambda$ arbitrarily small since we want to find the algorithmic threshold where $\Lambda=0$.  The term $Q'(m(t))$ becomes $Q''(0)m(t)$. Since $m(t)$ has an arbitrarily slow evolution, whereas $R_{\rm TTI}(t-t'')$ relaxes to zero on a finite timescale,
the second term of the RHS of eq. (\ref{eq:mag}) simplifies to:
\begin{equation}
	\delta \exp(\Lambda t)Q''(1)\int_{0}^t R_{\rm TTI}(t-t'') \exp(-\Lambda (t-t''))dt''\simeq m(t)Q''(1)\overline{R}  \nonumber
\end{equation}
where $\overline{R}=\int_{0}^t R_{\rm TTI}(t-t'')dt''$  does not depend on $t$ (since $t$ can be taken arbitrarily large and $R_{\rm TTI}(t-t'')$ relaxes to zero on finite time-scales).
The contribution of to the last term on (\ref{eq:mag}) reads:
\begin{equation}
	\begin{split}
		& \delta\exp(\Lambda t)\!\! \int_0^{t}\!\!\! R_{\rm ag}(t,t'')Q''\!(C_{\rm ag}(t,t'')) e^{\!\!-\Lambda (t- t'')} dt''
		\!\!=\!  m(t)\!\! \int_0^{t}\!\!\! R_{\rm ag}(t,t'')Q''\!(C_{\rm ag}(t,t''))e^{\!\!-\Lambda (t- t'')} ) dt''. \nonumber
	\end{split}
\end{equation}
Using that $Q''(C_{\rm ag}(t,t''))$ is bounded by $Q''(1)$ and that
$\Lambda$ cuts-off the integral on a time $t_0\sim 1/\Lambda$ that does not diverge with $t$, we can use the "weak-long term memory" property to conclude that the last term is arbitrarily small compared to $m(t)$ and hence can be neglected with respect to the previous ones. Collecting all the pieces together we find:
%
\begin{equation}
	\frac{d}{dt}m(t)= \left[-\mu_{\infty} +Q''(0)+ Q''(1)\overline{R}\right]m(t) + O(\delta^2)\,.
\end{equation}
This is solved by  $m(t)=\delta \exp(\Lambda t)$ with
$\Lambda=-\mu_{\infty} +Q''(0)+Q''(1)\overline{R} \,,$
which therefore justifies a posteriori our assumption of exponential growth. The condition for the instability of the aging solution towards the signal solution is therefore given by
\begin{equation}\label{eq:CHSCK_stability_condition}
	0=-\mu_{\infty} +Q''(0)+ Q''(1)\overline{R} \,.
\end{equation}
From the analysis of the asymptotic aging solution presented in SM Sec.~\ref{sec:CHSCK_CK_analysis} one finds
that $\mu_{\infty}=2\sqrt{Q''(1)}$ and  $\overline{R}=1/\sqrt{Q''(1)}$, therefore obtaining $Q''(0)=\sqrt{Q''(1)}$.
This condition is the same one found from the study of the landscape, and thus leads to the transition line eq.~(\ref{eq:GD_threshold_kac-rice}).


\subsubsection*{Acknowledgments}
We thank Pierfrancesco Urbani for many related discussions. 
	We acknowledge funding from the ERC under the European
	Union’s Horizon 2020 Research and Innovation Programme Grant
	Agreement 714608-SMiLe; from the French National
	Research Agency (ANR) grant PAIL; and from the Simons Foundation (\#454935, Giulio Biroli).


\medskip


\input{arxiv.bbl}


\appendix

\section{Kac-Rice method}\label{sec:kac-rice}

\subsection{Summary of the Kac-Rice complexity}
\label{sec:summary}
\begin{figure}[ht!]
	\centering
	\includegraphics[width=0.9\linewidth]{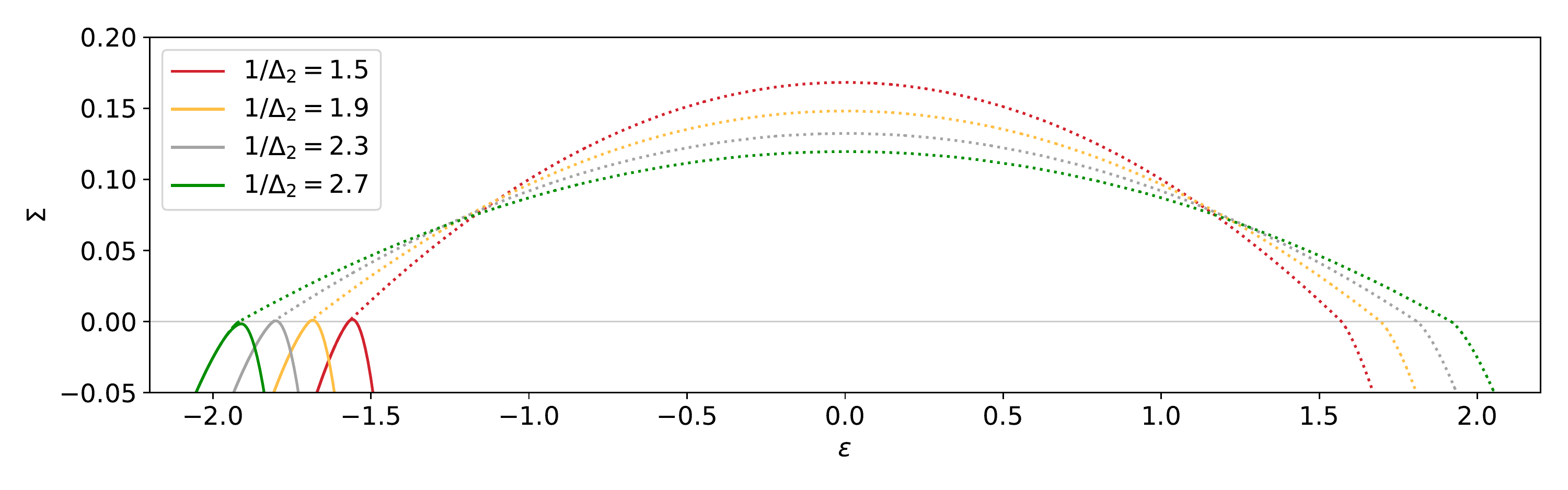}
	\subfigure[]{
	\includegraphics[scale=.402]{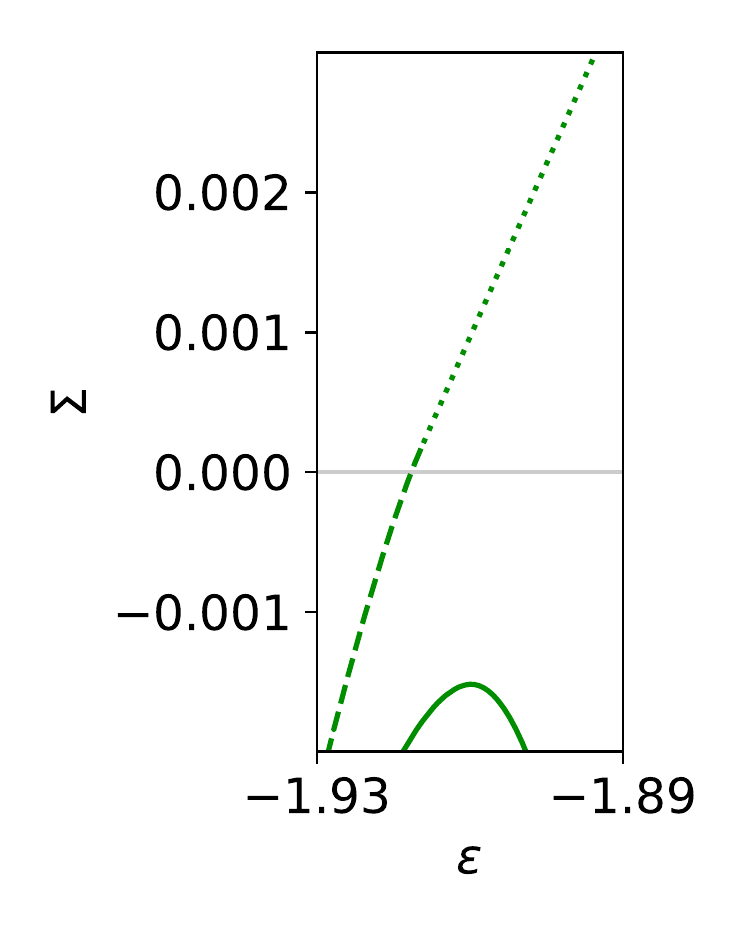}
	}
	\centering
	\subfigure[]{
	\includegraphics[scale=.402]{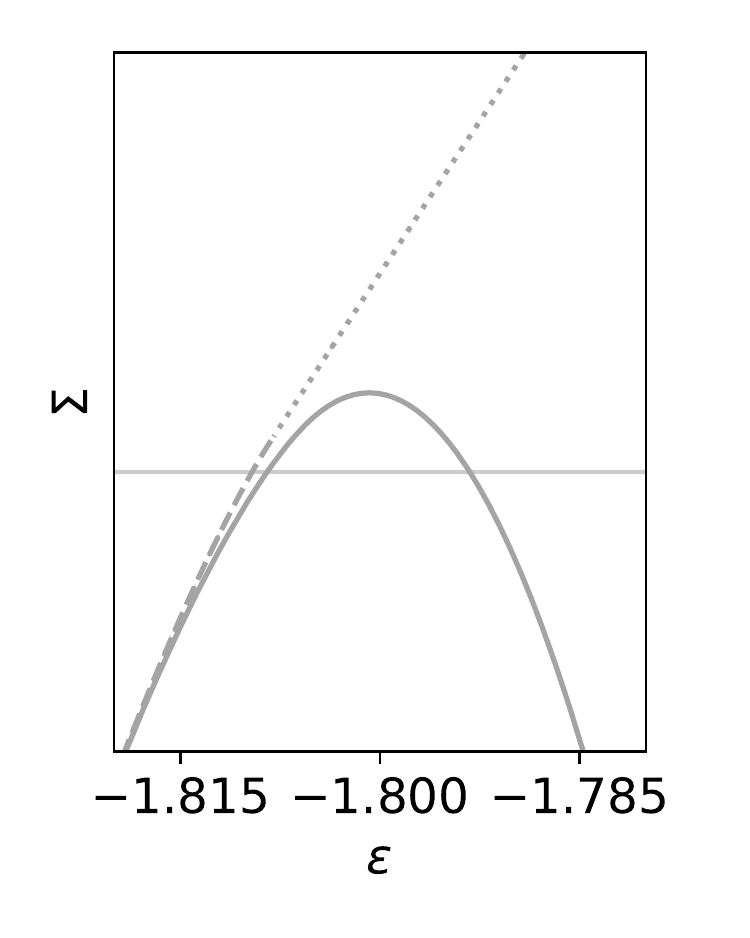}
	}
	\subfigure[]{
	\includegraphics[scale=.402]{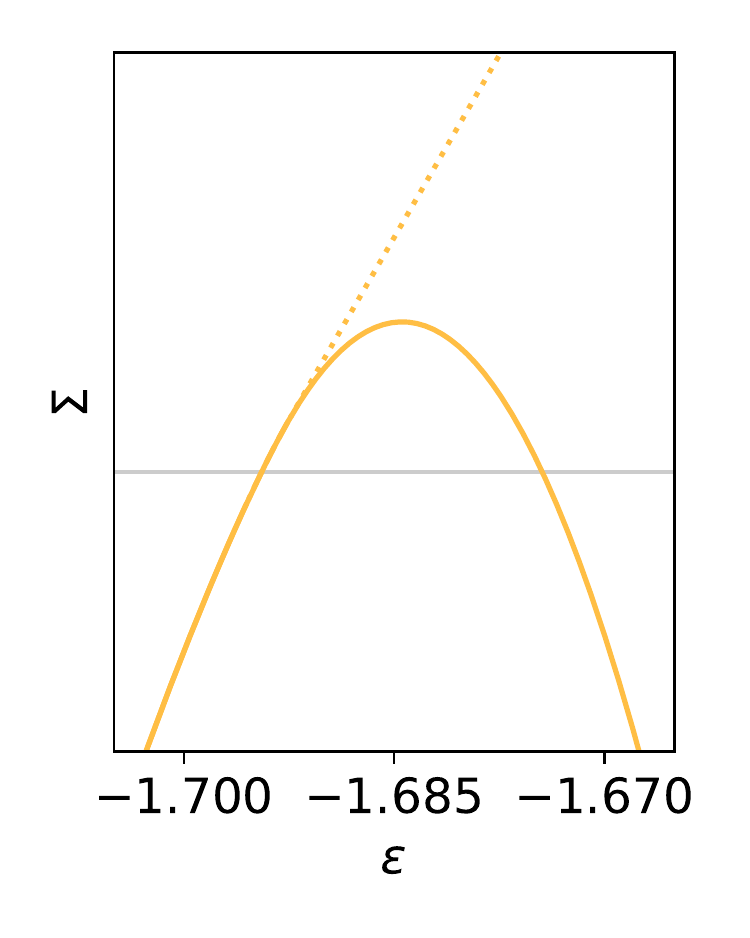}
	}
	\subfigure[]{
	\includegraphics[scale=.402]{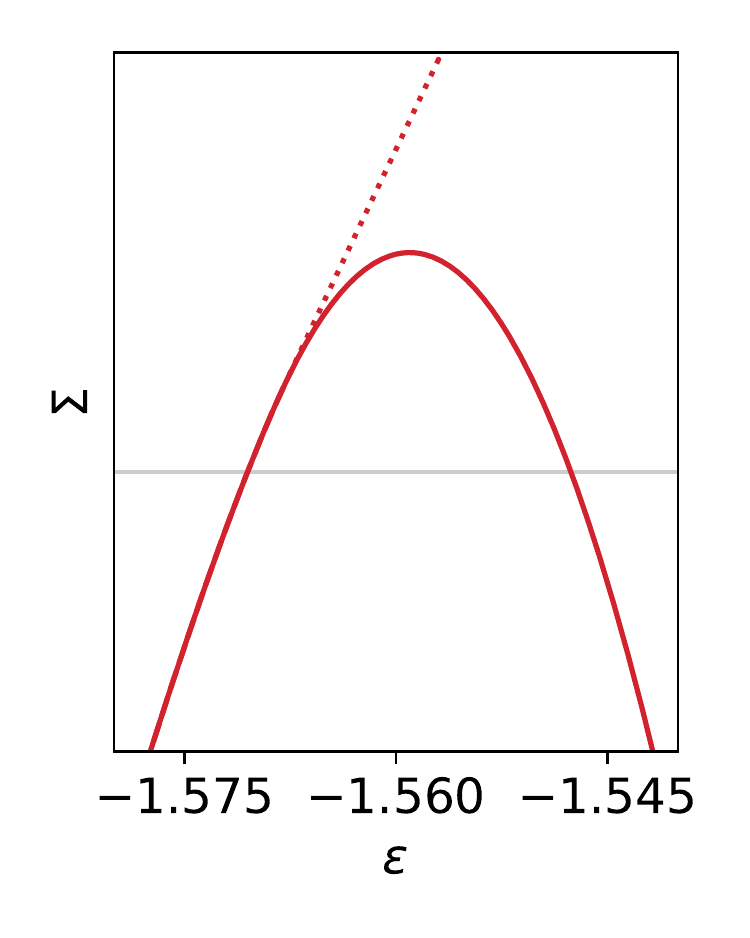}
	}
	\caption{Curves of the complexity of critical points, dotted and dashed curves from  Eq.~\eqref{eq:complexity_annealed_stationary}, and of minima, full curve from Eq.~\eqref{eq:complexity_annealed}, at overlap value $m=0$ at fixed $\Delta_p = 1.0$ for different $\Delta_2$. The figure shows qualitatively the same features as Fig.~\ref{fig:complexity_energy_zoom}, but displays the full positive part of the complexity for the four cases discussed in the main text, $1/\Delta_2 \in\{1.5,\,1.9,\,2.3,\,2.7\}$. 
		Zooms of the curves of the annealed complexity of critical points and minima when they cross zero at negative loss are in the panels labelled from (a) to (d) for increasing $1/\Delta_2$.}
	\label{fig:complexity_energy}
\end{figure}

In this section we introduce the Kac-Rice formula and we show how to reduce it to an explicit expression for the spiked matrix-tensor model.
The Kac-Rice formula evaluates the expected number of critical points of a rough function subject to a number of conditions.
For an inference problem it is interesting to focus on the expected number of critical points constrained to have of given loss and a given overlap with the ground truth. For convenience reasons
we consider the rescaled loss $\mathcal{L}(\pmb\sigma) = N\ell(\pmb\sigma)$. The Kac-Rice formula then reads
\begin{equation}
	\begin{split}
		\mathbb{E}_\eta [\mathcal{N}(\epsilon,m| \Theta)] &= \int_{\mathbb{S}^{N-1}} \delta\left(\langle\pmb\sigma,\pmb\sigma^*\rangle-m\right) \mathbb{E}_\eta\left[ |\det \mathcal H | \Big| \mathcal L = N\,\epsilon, \partial_i\mathcal L=0\ \forall i, \lambda_{\min}>0\right] \times
		\\
		&\times \phi_{ \mathcal L,\,\partial_i\mathcal L}(\pmb \sigma,\pmb 0,\epsilon) d\pmb \sigma\,,
	\end{split}
\end{equation}
where $\eta$ represents the noise in the problem, $\Theta$ the parameters and $\phi$ the joint probability density of the loss and its gradient.

The quantity of interest is
the density of the logarithm of the number of critical points $\log\mathcal{N}(\epsilon,m| \Theta)/N$.
It should be noted that, since the random variable representing the number of critical point fluctuates at the exponential scale, a correct estimation of the expected value of this quantity is not $\log\mathbb{E}_\eta [\mathcal{N}(\epsilon,m| \Theta)]$, as it would be immediately obtained by using the result of the Kac-Rice formula \cite{sarao2019passed}, but $\mathbb{E}_\eta [\log\mathcal{N}(\epsilon,m| \Theta)]$. These two quantities are called respectively \textit{annealed} and \textit{quenched complexities}.
Using Jensen inequality one observes that the annealed complexity is just an upper bound of the quenched one. However, for mathematical convenience most of the studies have been focused on the former. Eventually the second moment of the number of critical points has been evaluated \cite{subag2017complexity}, by an extension of the Kac-Rice formula to higher moments \cite{adler2009random}, just to prove that the two are equivalent in some models \cite{subag2017complexity}.
The quenched complexity has been evaluated in a related model in a non rigorous way by studying the $n$-th moment and applying replica trick, the so-called replicated Kac-Rice \cite{ros2018complex}. Given a random variable $Y$ replica trick says
\begin{equation}\label{eq:replica_trick}
	\mathbb{E}_\eta[\log Y] = \lim_{n\rightarrow0^+} \frac{\mathbb{E}_\eta[Y^n]-1}n
\end{equation}
but instead of considering an arbitrary $n\in\mathbb{R}^+$, the study is done using $n\in\mathbb{N}$ and performing an analytic continuation of the result to $0^+$. The replica trick has already been used in a plethora of applications and, although not rigorous, it was found correct in all naturally motivated cases that have been later approached by other techniques. An important mathematical literature has developed in order to understand the method.

In the next section we sketch the derivation of the quenched Kac-Rice and we provide all the information to determine the annealed one. Since the threshold is determined considering the configuration with arbitrarily small overlap $m\ll1$, we focus on that case. Remarkably we found that as $m\rightarrow0$ the quenched complexity computed within the replica symmetric (RS) approximation is equal to the annealed one\footnote{We expect the RS approximation to be correct for $p=3$ in the whole $(\Delta_p,1/\Delta_2)$ phase diagram, and hence 
that quenched and annealed complexities coincide. For $p>3$, results from replica theory \cite{CL04} suggest that one needs to go beyond the RS approximation at least in some parts of the phase diagram.}.
We show that the corresponding Hessian is Eq.~\eqref{eq:Hessian} in the main text, \textit{i.e.} it is proportional to a GOE translated by $t$ and perturbed by a rank $n$ perturbation of strength $\theta$ that in the annealed case is of rank $1$. Thus we find that the complexity for the stationary points is \cite{sarao2019passed}
\begin{equation}\label{eq:complexity_annealed_stationary}
	\begin{split}
		\Sigma_a^\text{sta}(m,\epsilon | \Delta_p,\Delta_2) 
		&= \max_{\substack{\epsilon_p,\epsilon_2\\ \text{s.t. } \epsilon_p + \epsilon_2 = \epsilon}} \frac12\log\frac{Q''(1)}{Q'(1)} + \frac12\log(1-m^2) -\frac12\frac{\left(Q''(m)\right)^2}{Q'(1)}(1-m^2) +
		\\
		& - \frac{p\Delta_p}2\left(\epsilon_p+\frac{m^p}{p\Delta_p}\right)^2 - \Delta_2\left(\epsilon_2+\frac{m^2}{2\Delta_2}\right)^2 + \Phi(t),
	\end{split}
\end{equation}
with
\begin{equation}\label{eq:Phi}
	\Phi(t) =  \begin{cases}
		 & \frac{t^2}4 \quad\quad \text{if } |t|\le2
		\\
		 & \frac{t^2}4 + \log\left(\sqrt{\frac{t^2}4-1}+\frac{|t|}2\right)-\frac{|t|}4\sqrt{t^2-4} \quad\quad \text{otherwise}
	\end{cases}
\end{equation}
and $t = -\left(p\epsilon_p+2\epsilon_2\right)/\sqrt{Q''(1)}$ as already introduced in the main text.
Finally studying the eigenvalue of the Hessian to constrain them in the positive semi-axis, we find the complexity of minima \cite{sarao2019passed}
\begin{equation}\label{eq:complexity_annealed}
	\Sigma_a(m,\epsilon | \Delta_p,\Delta_2) =  \Sigma_a^\text{sta}(m,\epsilon_p,\epsilon_2 | \Delta_p,\Delta_2) - L(\theta,t)\,.
\end{equation}
with
\begin{equation}\label{eq:large_deviation_function_lambda_min}
	L(\theta,t) = \begin{cases}
		 & \frac14\int_{\theta+\frac1\theta}^t \sqrt{y^2-4}dy -\frac\theta2\left(t-\left(\theta+\frac1\theta\right)\right)
		+\frac{t^2-\left(\theta+\frac1\theta\right)^2}8 \quad \theta>1,\; 2\le t<\frac{\theta^2+1}\theta
		\\
		 & \infty \quad\quad t<2                                                                                           \\
		 & 0 \quad\quad \text{otherwise.}                                                                                  \\
	\end{cases}
\end{equation}
and $\theta = Q''(m)\,(1-m^2)/\sqrt{Q''(1)}$. In Fig.~\ref{fig:complexity_energy}, we show the two complexities of the stationary points and of the minima in the parameter space discussed in the main text with discontinuous lines Eq.~\eqref{eq:complexity_annealed_stationary} and full lines Eq.~\eqref{eq:complexity_annealed}, respectively. A positive complexity means an exponential number of critical points (minima). The region where exponentially many minima appear is highlighted in the small figures, showing the coexistence of exponentially many minima and saddles.
\subsection{Derivation of the quenched complexity}\label{sec:kac-rice_derivation}
We proceed with the computation of the quenched Kac-Rice complexity for the spiked matrix-tensor model, using replicated Kac-Rice prescription for the spiked pure-tensor model \cite{ros2018complex}. This implies, following replica trick Eq.~\eqref{eq:replica_trick}, the evaluation of the $n$-th moment of number of minima using Kac-Rice formula which is given by \cite{adler2009random}
\begin{equation}
	\begin{split}
		\mathbb{E}_\eta [\mathcal{N}(\epsilon,m| \Theta)^n] \! &= \hspace{-.2cm}\int_{\mathbb{S}^{N-1}} \hspace{-.5cm}\cdots \int_{\mathbb{S}^{N-1}}\hspace{-.4cm} \mathbb{E}_\eta\left[ \left(\prod_{a = 1}^n\left| \det  \mathcal H[\pmb\sigma^a] \right| \right)\Bigg| \forall b,c\ \mathcal L[\pmb\sigma^b] = N\epsilon, \partial_i\mathcal L[\pmb\sigma^c]=0\ \forall i, \lambda_{\min}\!\!>0\right]
		\\
		& \times \phi_{ \mathcal L,\,\partial_i\mathcal L} \left(\{\pmb\sigma^a\},\pmb 0,\epsilon \right) \prod_{a=1}^n \delta\left(\langle\pmb\sigma^a,\pmb\sigma^*\rangle-m\right) d\pmb \sigma^a\,.
	\end{split}
\end{equation}
Where $\phi$ is the joint probability density of the loss and its gradients evaluated on the $n$ replicated configurations. We hereby sketch the main computation steps and present the results that are the most relevant for the theory presented in this paper, i.e. for $m=0$. The details of the computation and further results for $m>0$ will be presented in a dedicated work elsewhere.

It is convenient to consider free variables on $\mathbb{R}^N$ and constrain them using a Lagrange multiplier~$\gamma$. Thus $\mathcal{L}(\pmb\sigma^a)\mapsto \mathcal{L}(\pmb\sigma^a) - \frac\gamma2\left(\sum_i (\sigma^a_i)^2 -1\right)$. Using the fact that the gradient must be zero on the sphere, {\it i.e.} that
$\nabla_i\mathcal{L}(\pmb\sigma^a) - \gamma\,\sigma^a_i = 0 \ \ \forall i$, we obtain a simple expression for the multiplier:
$\gamma=\langle \pmb\nabla\mathcal{L}(\pmb\sigma^a), \pmb\sigma^a\rangle$.
Moreover by separating the two terms in the loss that represent the contribution of the two channels, $\mathcal{L}=\mathcal{L}_p+\mathcal{L}_2$, we define $\mathcal{L}_p=N\epsilon_p$ and $\mathcal{L}_2=N\epsilon_2$ so to obtain for the multiplier the even simpler equation $\gamma = p\epsilon_p  + 2\epsilon_2$.
To take advantage of this simple formula, in the following we work with the contributions of the two channels to the loss function separately and we impose the constraint on their sum, $N \epsilon = N (\epsilon_p  + \epsilon_2)$, only at the end. 

The use of Cartesian coordinates allows us to evaluate easily the moments and covariances by means of standard derivatives.
\begin{align}
	\label{eq:mean_loss}
	 & \mathbb{E}_\eta\left[\mathcal{L}[\pmb\sigma^a]\right] = - N Q(\langle \pmb\sigma^a,\pmb\sigma^*\rangle)\, ,
	\\
	\label{eq:covariance_loss}
	 & \Cov\left[\mathcal{L}[\pmb \sigma^a],\mathcal{L}[\pmb \sigma^b]\right] = N\,Q\left(\langle \pmb\sigma^a, \pmb\sigma^b\rangle\right)\, .
\end{align}
Taking derivatives of these equation gives all the covariances of loss, gradient and Hessian. For instance, we can easily see that the covariance of the Hessian is given by:
\begin{equation}
	\begin{split}
		& \frac{\partial^4}{\partial\sigma^a_i\partial\sigma^a_j \partial\sigma^b_k \partial\sigma^b_l} \frac{\Cov\left[\mathcal{L}[\pmb \sigma^a], \mathcal{L}[\pmb \sigma^b]\right]}N = Q''''\left(\langle \pmb\sigma^a, \pmb\sigma^b\rangle\right)\langle \pmb\sigma^b, \pmb e_i^a\rangle \langle \pmb\sigma^b,\pmb e_j^a\rangle \langle \pmb\sigma^a,\pmb e_k^b\rangle \langle \pmb\sigma^a, \pmb e_l^b\rangle +
		\\
		& \,+Q'''\left(\langle \pmb\sigma^a, \pmb\sigma^b\rangle\right)\! \Big(\langle \pmb e_i^a, \pmb e_k^b \rangle \langle \pmb\sigma^b, \pmb e_j^a\rangle \langle \pmb\sigma^a, \pmb e_l^b\rangle \!+\! \langle \pmb e_j^a, \pmb e_k^b\rangle \langle \pmb\sigma^b, \pmb e_i^a\rangle \langle \pmb\sigma^a,\pmb e_l^b\rangle \!+\! \langle \pmb e_i^a,\pmb e_l^b\rangle \langle \pmb\sigma^b, \pmb e_j^a\rangle \langle \pmb\sigma^a,\pmb e_k^b\rangle +
		\\
		& \,+\langle \pmb e_j^a, \pmb e_l^b\rangle \langle \pmb\sigma^b,\pmb e_i^a\rangle \langle \pmb\sigma^a, \pmb e_k^b\rangle \Big) + Q''\left(\langle \pmb\sigma^a, \pmb\sigma^b\rangle\right) \Big(\langle \pmb e_i^a, \pmb e_k^b\rangle \langle \pmb e_j^a, \pmb e_l^b\rangle +\langle \pmb e_i^a, \pmb e_l^b\rangle \langle \pmb e_j^a, \pmb e_k^b\rangle \Big)\,,
	\end{split}
\end{equation}
where $\{\pmb e^a_i\}_i$ and $\{\pmb e^b_k\}_k$ are the reference frames associated to replica $a$ and $b$ respectively.

\begin{remark}[annealed Hessian]
	In particular notice that if $n=1$ there is only one replica and using an orthogonal basis where the $N$-th direction is aligned with the replica and projecting on the sphere by discarding the last coordinates we obtain a simple expression:
	\begin{equation}\label{eq:covariance_hessian_annealed}
		\frac1N\Cov\left[\mathcal{H}_{ij}[\pmb\sigma],\mathcal{H}_{kl}[\pmb\sigma]\right] = Q''(1)\left(\delta_{ik}\delta_{jl} + \delta_{il}\delta_{jk}\right)
	\end{equation}
	with the $delta$ representing Kronecker's deltas. This is the expression of a GOE. We can as well compute the mean deriving twice in the $i$-th and $j$-th coordinate. Following \cite{arous2017landscape} we make another convenient choice for the basis imposing that the signal lies in the space spanned by the $\pmb e_1$ and $\pmb e_N = \pmb\sigma$. This gives,
	\begin{equation}
		\frac1N\mathbb{E}\left[\mathcal{H}_{ij}[\pmb\sigma]\right] = Q''(\langle \pmb\sigma^a,\pmb\sigma^*\rangle)\,\langle\pmb\sigma^*,\pmb e_i\rangle\langle\pmb\sigma^*,\pmb e_j\rangle = Q''(\langle \pmb\sigma^a,\pmb\sigma^*\rangle)\,\langle\pmb\sigma^*,\pmb e_i\rangle\langle\pmb\sigma^*,\pmb e_j\rangle \delta_{i1}\delta_{j1}
	\end{equation}
	that, when $m=0$, equals
	\begin{equation}\label{eq:mean_hessian_annealed}
		\frac1N\mathbb{E}\left[\mathcal{H}[\pmb\sigma]\right] = Q''(0)\,\pmb e_1 \pmb e_1^T\,.
	\end{equation}
	Wrapping together Eq.~\eqref{eq:covariance_hessian_annealed}, Eq.~\eqref{eq:mean_hessian_annealed} and the expression for the Langrange multiplier that acts as a translation, we obtain the Hessian presented in the main text Eq.~\eqref{eq:Hessian}. Observe, however, that the Hessian in which we are interested in is not the simple Hessian of the loss but rather the Hessian of the loss conditioned to a given loss and a given gradient. Using Eq.~\eqref{eq:covariance_loss} to compute the covariance of Hessian and loss, and of Hessian and gradient under this basis, we can observe that these random variables are unconditioned. Thus the conditioning does not affect the distribution of the Hessian of the loss and therefore Eq.~\eqref{eq:Hessian} is recovered. 
\end{remark}

Eq.~\eqref{eq:mean_loss} and Eq.~\eqref{eq:covariance_loss} are basic ingredients required to continue with the analysis. In the next two sections we first compute the joint density of the loss and its gradient, and second compute the expected value of the determinant of the Hessians. In the final section we put together the results obtaining the complexities already presented in the summary \ref{sec:summary}.

\subsubsection{Joint probability density.}\label{sec:kac-rice_joint_density} In order to evaluate the joint probability density $\phi$ we focus on the covariance matrix of the loss and its gradient, that using Eq.~\eqref{eq:covariance_loss} is given by:
\begin{equation}
	\label{eq:covariance_grad_loss_matrix}
	\frac{1}{N}\left[\pmb C_{\mathcal{L},\nabla\mathcal{L}}\right]^{a,b} = \begin{bmatrix}
		Q''\left(\langle \pmb\sigma^a, \pmb\sigma^b\rangle\right)\pmb\sigma^a \otimes \pmb\sigma^b + Q'\left(\langle \pmb\sigma^a, \pmb\sigma^b\rangle\right) \mathbb{I}_N & Q'\left(\langle \pmb\sigma^a, \pmb\sigma^b\rangle\right) \pmb\sigma^{b,T} \\
		Q'\left(\langle \pmb\sigma^a, \pmb\sigma^b\rangle\right) \pmb\sigma^a                                                                                              & Q\left(\langle \pmb\sigma^a, \pmb\sigma^b\rangle\right)
	\end{bmatrix}\,.
\end{equation}
The joint density corresponds to the probability of observing a zero gradient on the sphere and a given loss, $(\gamma\,\pmb\sigma^T, \epsilon)^T$, in the multivariate Gaussian variable $(\pmb\nabla\mathcal{L}^T, \mathcal{L})^T$. Thus taking into account the first moments of loss and gradient, obtained from Eq.~\eqref{eq:mean_loss}, we define the auxiliary vector 
\begin{equation}
[\pmb\mu(\epsilon_p,\epsilon_2)]^a = \left((p\epsilon_p+2\epsilon_2)\pmb\sigma^{a,T} + Q'(\langle \pmb\sigma^a,\pmb\sigma^*\rangle)\pmb\sigma^{*,T},\ \epsilon + Q(\langle \pmb\sigma^a,\pmb\sigma^*\rangle)\right)^T\, .
\end{equation} 
The probability density is given by
\begin{equation}
	\phi_{ \mathcal L,\,\partial_i\mathcal L} \left(\{\pmb\sigma^a\},\pmb 0,\epsilon \right) \!\propto\!\! \iint\!\!\delta(\epsilon-\epsilon_p-\epsilon_2) \exp\!\!\left[-\frac12\sum_{a,b}\
	[\pmb\mu(\epsilon_p,\epsilon_2)]^{a,T}
	\left[\pmb C_{\mathcal{L},\nabla\mathcal{L}}^{-1}\right]^{a,b}\!
	[\pmb\mu(\epsilon_p,\epsilon_2)]^b \right]
	  d\epsilon_p d\epsilon_2.
	\label{eq:integrand_density}
\end{equation}
This expression can be evaluated by observing that there is a set of $(N+1) n$-dimensional vectors that forms a closed group under the action of the covariance matrix Eq.~\eqref{eq:covariance_grad_loss_matrix}.
This set is composed by the following four vectors
\begin{align}
	 & \pmb\xi_1^T = \left(\pmb\sigma^{1,T}, 0, \pmb\sigma^{2,T}, 0, \dots,\pmb\sigma^{n,T}, 0\right)\, ,
	\\
	 & \pmb\xi_2^T = \left(
	\sum_{e\neq1}\pmb\sigma^{e,T}, 0,
	\sum_{e\neq2}\pmb\sigma^{e,T}, 0,
	\dots,
	\sum_{e\neq n}\pmb\sigma^{e,T}, 0\right)\, ,
	\\
	 & \pmb\xi_3^T = \left(\pmb 0^T, 1, \pmb 0^T, 1, \dots,\pmb 0^T, 1\right)\, , 
	\\
	 & \pmb\xi_4^T = \left(\pmb\sigma^{*,T}, 0, \pmb\sigma^{*,T}, 0, \dots,\pmb\sigma^{*,T}, 0\right) \ ,
\end{align}
where $\pmb 0$ is an $N$ dimensional null vector. Indeed the auxiliary vector can be rewritten in terms of the elements of this set of newly defined vectors as follows
\begin{equation}
	[\pmb\mu(\epsilon_p,\epsilon_2)]^a = (p\epsilon_p+2\epsilon_2) [\pmb\xi_1]^a + (\epsilon + Q(\langle \pmb\sigma^a,\pmb\sigma^*\rangle)\,[\pmb\xi_3]^a + Q'(\langle \pmb\sigma^a,\pmb\sigma^*\rangle)\pmb\,[\xi_4]^a\,.
\end{equation}
At this point we exploit the fact that the set of these vectors forms a closed group under the action of the covariance matrix. In fact we can invert its action on the set $\{\pmb\xi_k\}_{k=1}^4$ only, without the need to evaluate the inverse of the full covariance matrix.
Using this trick, the integrand in Eq.~\eqref{eq:integrand_density} can be evaluated.
The result for the integrand in  Eq.~\eqref{eq:integrand_density} contains the dependence on the configurations of replicas only in terms of the overlaps $q_{a,b}=\langle\pmb \sigma^a, \pmb \sigma^b\rangle$ with each other, and of the overlap of each of them with the ground truth, {\it i.e.} the magnetisation $m_a=\langle \pmb \sigma^a, \pmb \sigma^*\rangle$.
In this formulation, hence, the problem of evaluating a free integral over $n$ vectors on the sphere has been translated into the task of evaluating an integral over the possible choices of the $n \times n$ matrix of the overlaps
provided that we consider the multiplying factor that comes from the volume $V(\{q_{a,b}\},\{m_a\})$ of configurations that are compatible with that choice and the condition on the magnetisations. 

The next step is to make an ansatz on the form of the matrix of these overlaps which must be consistent with the condition on the vector of magnetisations required in the Kac-Rice formula. The simplest ansatz is called {\it replica symmetric} ansatz and assumes that the overlaps of different replicas are independent of the indices $a$ and $b$, i.e.
\begin{equation}
	\langle\pmb\sigma^a, \pmb\sigma^b\rangle = \delta_{ab} + (1-\delta_{ab})\,q\,.
\end{equation}
Note that the replica symmetric ansatz is compatible with the condition $\langle\pmb\sigma^a, \pmb\sigma^*\rangle = m\ \forall a$ imposed in the Kac-Rice formula.
Within this ansatz the probability density can be evaluated as a function of $q$ and $m$ for arbitrary $n$ and the analytic continuation at $n\rightarrow0^+$ can be finally taken to evaluate the quenched complexity.
The expression for generic $n$ is too long and convoluted to be reported here.
However in the limit $n\rightarrow 1$ it corresponds to the expression of the probability density of losses and gradients evaluated in the annealed computation which has the following nice expression
\begin{equation}
	\begin{split}
		& \phi_{ \mathcal L,\,\partial_i\mathcal L}\left(\{\pmb\sigma^a\},\pmb 0,\epsilon \right) \propto \int\int \delta(\epsilon-\epsilon_p-\epsilon_2) \exp\left[ -\frac{N}2\frac{\left(Q''(m)\right)^2}{Q'(1)}(1-m^2)\right]\times
		\\
		&\times\exp\left[
			- \frac{Np}2\Delta_p\left(\epsilon_p + \frac{m^p}{p\Delta_p}\right)^2 - N\Delta_2\left(\epsilon_2 + \frac{m^2}{2\Delta_2}\right)^2\right]\,d\epsilon_p d\epsilon_2
		\\
		&\simeq \max_{\substack{\epsilon_p,\epsilon_2 \\\text{s.t. }\epsilon_p + \epsilon_2 = \epsilon}} \exp\left[ -\frac{N}2\frac{\left(Q''(m)\right)^2}{Q'(1)}(1-m^2)
			- \frac{Np}2\Delta_p\left(\epsilon_p + \frac{m^p}{p\Delta_p}\right)^2 - N\Delta_2\left(\epsilon_2 + \frac{m^2}{2\Delta_2}\right)^2\right]
		.
	\end{split}
\end{equation}

We must also consider the normalisation of the density that is given by
\begin{equation}\label{eq:normalisation_density}
    \exp\left[-\frac{Nn}2\log(2\pi Q'(1))\right]\,.
\end{equation}

Finally we come back to the volume term $V(\{q_{a,b}\},\{m_a\})$.
Constraining the configurations to a given overlap $q$ with each other and $m$ with the ground truth produces a volume term that can be easily evaluated as
\begin{equation}
	V(q, m) \simeq \exp\left[\frac{Nn}2\left(\log\frac{2\pi e(1-q)}N-\frac{m^2-q}{1-q}\right)\right]\,,
\end{equation}
and for one single replica (which is useful in the computation of the annealed complexity) is simply
\begin{equation}
	V(m) \simeq \exp\left[\frac{N}2\left(\log\frac{2\pi e}{N}+\log(1-m^2)\right)\right]\,.
\end{equation}
Under the replica symmetric assumption we make a Laplace approximation that allows to evaluate the quenched complexity as an extremisation of the entire expression that depends on the overlap variable $q$ through the volume term $V(q,m)$ and the probability density $\phi$.
An interesting remark concerns the limit $q\rightarrow0$ in quenched joint probability density. Indeed in that case the two joint probability coincide module a factor $n$. We checked numerically that as $m\rightarrow0$ the optimal $q$ goes to $0$, which implies that the equations of the annealed and quenched complexities do correspond on the equator $m=0$.

\subsubsection{Expected value of the Hessian.}\label{sec:kac-rice_hessian}

As discussed introducing Eq.~\eqref{eq:covariance_loss}, the Hessian is a matrix-valued random variable with multivariate Gaussian distribution.
In evaluating the Kac-Rice formula we must consider the distribution of the Hessian conditioned to the loss and its gradient, this can be imposed using the formula for conditioning of Gaussian random variables. Given $\pmb X,\pmb Y$ Gaussian random variables with covariance $\pmb C$ and mean $\pmb \mu$ the distribution of $\pmb X$ conditioned to $\pmb  Y=\pmb y^*$ is still Gaussian with covariance and mean
\begin{align*}
	 & \pmb C_{\pmb X| \pmb Y = \pmb y^*} = \pmb C_{\pmb X} - \pmb C_{\pmb{XY}}\pmb C_{\pmb Y}^{-1}\pmb C_{\pmb{XY}}\,;               \\
	 & \pmb \mu_{\pmb X|\pmb Y = \pmb y^*} = \pmb \mu_{\pmb X} + \pmb C_{\pmb{XY}}\pmb C_{\pmb Y}^{-1}(\pmb y^*-\pmb \mu_{\pmb Y})\,.
\end{align*}
In the annealed case, by using Eq.~\eqref{eq:covariance_loss} and the expression for the Langrange multiplier $\gamma$ we get that the Hessian corresponds to a shifted GOE subject to a rank one perturbation, as already discussed in the main text (see Eq.~\eqref{eq:Hessian}).
\begin{figure}[ht!]
	\centering
	\includegraphics[width=0.5\linewidth]{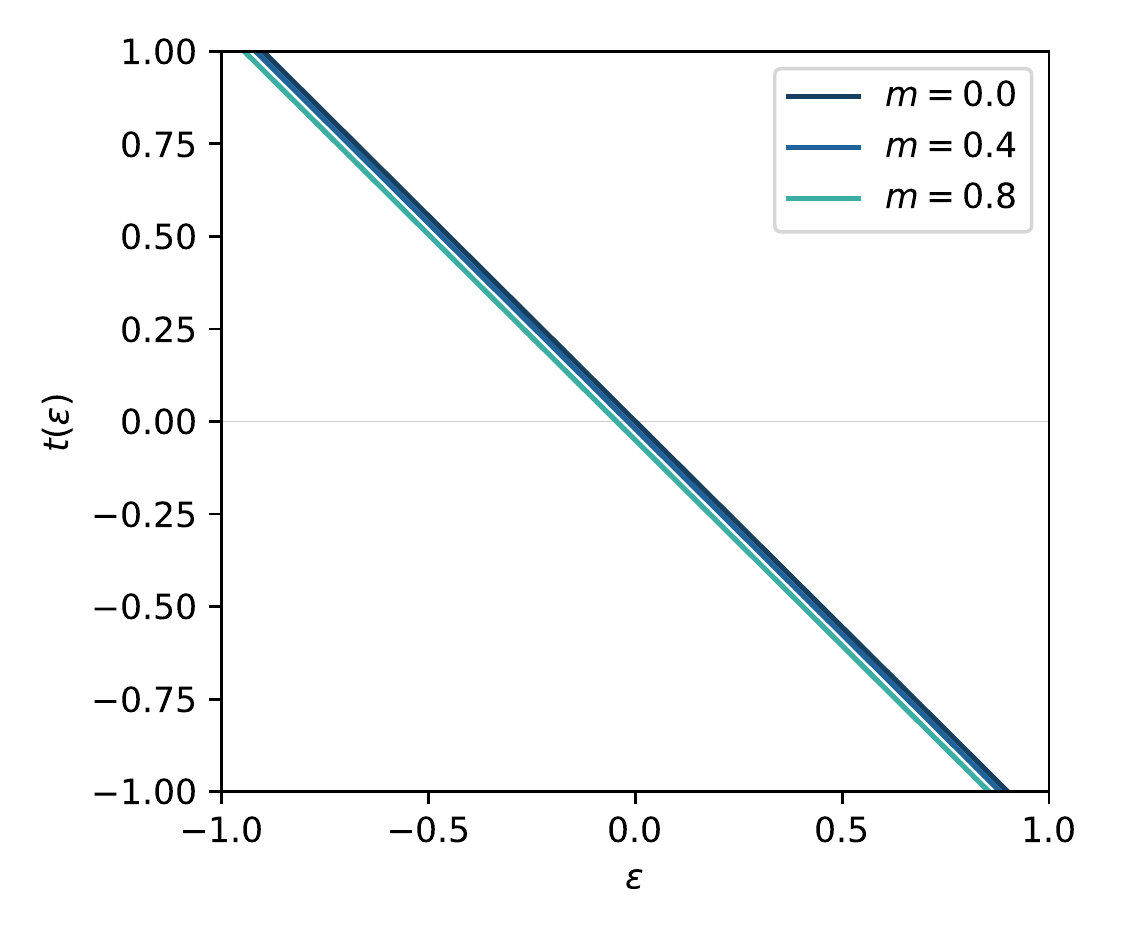}
	\caption{Shift of the Hessian, from Eq.~\eqref{eq:Hessian}, as function the loss density for different values of $m$. The qualitative behaviour shown in the figure does not change varying the parameter of the systems, i.e. it is always a decreasing function. The figure shows results obtained using $p=3$, $\Delta_p=1.0$ and $1/\Delta_2=2.3$.}
	\label{fig:t_energy}
\end{figure}
In the replicated Kac-Rice formula a more complicated expression appears that depends on the product of the determinants of Hessians associated to different replicas.
However,
using a proper reference frame, it was already noticed \cite{subag2017complexity,ros2018complex} that each Hessian corresponds also to a GOE since it is dominated by a $(N-n) \times (N-n)$ GOE block as $n\ll N$.
Moreover it has also been shown \cite{subag2017complexity,ros2018complex} that the expectation value of the product the Hessian determinants is equivalent to the product of the expectation values of each determinant. 
Thus we can still use standard results on the spectrum of GOE random matrices to evaluate the term in the Kac-Rice that depends on the Hessian. The distribution of the spectrum of the eigenvalue is given by
\begin{equation}
    \rho(\lambda)d\lambda = \frac{\sqrt{4 Q''(1) - (\lambda + \gamma)^2}}{2\pi Q''(1)} d\lambda\,,
\end{equation}
thus the determinant is given by
\begin{equation}
    \int \rho(\lambda) \log|\lambda|\, d\lambda = \frac12 \log[2 Q''(1)] + \frac1\pi \int \sqrt{2-\lambda^2}\log\left|\lambda - \gamma/\sqrt{Q''(1)} \right| d\lambda\,.
\end{equation}
where we recognise $t = \gamma/\sqrt{Q''(1)}$.
After some algebra we find:
\begin{equation}
     \frac12 \log[Q''(1)] - \frac12 + \Phi(t)
\end{equation}
with $\Phi(t)$ defined in Eq.~\eqref{eq:Phi}.

\subsubsection{Complexities: Putting pieces together}\label{sec:kac-rice_complexities}

By putting the above pieces together we obtain the annealed complexity of stationary points Eq.~\eqref{eq:complexity_annealed_stationary} where we can finally distinguish the origin of the various terms: the first term comes from the normalisation of the density and the determinant of the Hessian, the second comes from the volume prefactor, the third, fourth and fifth terms are originated by the probability density of loss and gradient and the last term comes from the product of Hessians. 

In order to select only the minima in the study of the complexity we must impose that the smallest eigenvalue is positive. There are two possible scenarios: either the smallest eigenvalue is determined by the left edge of the bulk of the spectrum (the perturbation does not induces any BBP transition), or it is outside the bulk of the spectrum.
In the first case the probability that the smallest eigenvalue is positive is suppressed by a factor $e^{-N^2}$ and the corresponding large deviation function is infinite.
In the second case the large deviation function associated to the shift in the position of the smallest eigenvalue, that would allow to keep it positive, is finite and must be evaluated.
The problem can be addressed with a replica computation \cite{ros2018complex} and focuses only on the typical value of the eigenvalue, missing the large deviation function.
As we already discussed as $m\rightarrow0$ we found numerically that the overlap $q$ that extremises the complexity is $q=0$, which leads back to the annealed complexity as we have shown computing the density.
Since the main focus in the paper is on the critical points at the equator we do not compute the isolated eigenvalue in a quenched approach but we rather use the large deviation function as it can be obtained in the annealed approximation \cite{arous2018algorithmic} of which we report here the result.
The condition on having a positive minimum eigenvalue suppresses the number of critical points by a factor $e^{- N L(\theta,t)}$, with $L(\theta,t)$ given in Eq.~\eqref{eq:large_deviation_function_lambda_min}, that enters in Eq.~\eqref{eq:complexity_annealed_stationary} leading to Eq.~\eqref{eq:complexity_annealed}.

\begin{remark}[threshold loss]
    Considering Eq.~\eqref{eq:complexity_annealed_stationary} and changing variables from $(\epsilon_p,\epsilon_2)$ to $(\epsilon,t)$ in the maximisation. We can now specialise on the region with no overlap with the signal, $m=0$, and consider that the Hessian of the loss touches the zero, $t=2$ which simplifies the expression of the piecewise term $\Phi(t)$. The result is a quadratic equation in $t$ which is equal to $2$ because of the marginality condition and can be solved in the energy $\epsilon$. The resulting energy is the Kac-Rice threshold energy, 
    \begin{equation}\label{eq:threshold_energy_KR}
        \epsilon_\text{\rm th}^\text{\rm KR} = \frac1{\sqrt{Q''(1)} Q'(1)}\left[\frac{(p-2)^2}{2p\Delta_2\Delta_p}-2 Q''(1) Q(1)\right]\,.
    \end{equation}
\end{remark}

\newpage

\section{CHSCK Equations}\label{sec:CHSCK_appendix}

In this section we present a derivation of CHSCK equations for the spiked matrix-tensor model using the generating functional formalism and later the asymptotic analysis under the hypothesis presented in the main text.
The starting point is the loss, Eq.~\eqref{eq:loss}, expliciting the observations, Eqs.~(\ref{eq:observationT}-\ref{eq:observationY}),

\begin{equation}
	\begin{split}
		\loss(\pmb\sigma | \pmb T, \pmb Y) &= -\frac{\sqrt{(p-1)!}}{\Delta_p\sqrt{N}}\sum_{i_1<\dots<i_p} \eta_{i_1\dots i_p} \sigma_{i_1}\dots\sigma_{i_p} - \frac1{p\Delta_p}\langle \pmb \sigma, \pmb \sigma^*\rangle^p + \\
		& - \frac1{\Delta_2\sqrt{N}}\sum_{i<j} \eta_{ij} \sigma_i \sigma_j  - \frac1{2\Delta_2}\langle \pmb \sigma, \pmb \sigma^*\rangle^2
	\end{split}
\end{equation}
and the gradient flow Eq.~\eqref{eq:gradient_flow} that for mathematical convenience we associate to an auxiliary function $\pmb f(\pmb\sigma)$
\begin{equation}
	\dot\sigma_i(t) = -\mu(t)\sigma(t) - \frac{\partial \loss(\pmb\sigma(t) | \pmb T, \pmb Y)}{\partial \sigma_i(t)} \doteq f_i\left(\pmb\sigma(t)\right)\,.
\end{equation}

The next section shows in detail the derivation that proceeds by introducing a probability distribution for the different evolutions, or trajectories, of the dynamics at a fixed realization of the noise. Then the distribution is averaged over the noise and this implies some technical steps before obtaining the final form. The resulting distribution is used to average correlation, response function and magnetisation over all the trajectories giving the CHSCK Eqs.~(\ref{eq:CHSCK_C}-\ref{eq:CHSCK_mu}). In the analysis an important role is played by the normalisation constant of the distribution of the trajectories, that is used in the final steps to derive with simplicity the equations.

After deriving the equations we show how to apply the hypothesis on the large time behaviour of $t$ and $t'$ to the CHSCK Eqs. In the last part this analysis provides the constants $\overline{R}$ and $\mu_\infty$ used to the derive the threshold in the main text, and some interesting additional information, such as the value of the loss at the threshold shown in the right panel of Fig.~\ref{fig:dynamics_complexity}.

\subsection{Derivation of CHSCK Equations}\label{sec:CHSCK_derivation}

The first step is to discretise the time in $M$ time steps of length $h$. 
We want the trajectories to be a solution at every time step of Eq.~\eqref{eq:gradient_flow}, which discretized looks as $\sigma_i^{a+1} - \sigma_i^a = f_i\left(\pmb\sigma^a\right)h$ with $a$ the time index. Let's call $y^{a+1}$ a solution to this equation. We can define the probability density of observing a trajectory satisfying Eq.~\eqref{eq:gradient_flow} at a fixed noise:
\begin{equation}
	p(\pmb\sigma^1,\dots,\pmb\sigma^M) = \int \prod_{ai} \delta\left(\sigma_i^{a+1} - y_i^a(\pmb\sigma^a)\right) \prod_{a = 0}^{M-1} d\mu_{\mathbb{S}}^a\,.
\end{equation}
where $\mu_{\mathbb{S}}$ is the measure over $\mathbb{S}^{N-1}$.

The normalisation constant is the integral of this probability and is called {\it generating functional} $\mathcal{Z}$ and since the previous object is already properly normalised it is equal to $1$. Rewriting the $\delta$s as Fourier transforms and therefore including the auxiliary variables $\tilde{\pmb \sigma}^a$,
\begin{equation}
	1 = \int \prod_{ai}\exp\left[N\tilde{\sigma}_i^a\left(\sigma_i^{a+1} - \sigma_i^a - f_i(\pmb \sigma^a)h\right)\right] \prod_{a=0}^{M-1} d\mu_{\mathbb{S}}^a \frac{d\tilde{\pmb \sigma}^a}{2\pi i}
\end{equation}
where in order to have mathematically well-defined quantities in the large $N$ limit we have a factor in the exponential. Moving to the continuum, the generating functional appears as a path integral
\begin{equation}\label{eq:generating_functional}
	1 = \mathcal{Z} = \int \mathcal{D}\left[\pmb\sigma,\tilde{\pmb\sigma}\right] \prod_{i}\exp\left[N\int \tilde{\sigma}_i(t)\left(\partial_t\sigma_i(t) - f_i(\pmb \sigma(t))\right)dt\right]\,.
\end{equation}
So far the object we derived is a distribution that tells whether a trajectory from arbitrary initial condition respects or not gradient-flow dynamics, however, our interest is in average trajectories with respect to the realization of the disorder. Therefore the distribution has to be averaged and after some algebraic manipulation gives the average generating functional in Eq.~\eqref{eq:averaged_generating_functional},
%
\begin{equation}
	\begin{split}
		1 = \mathbb{E}_\eta &\left[\mathcal{Z}\right] = \int \mathcal{D}\left[\pmb\sigma,\tilde{\pmb\sigma}\right] \prod_{i}\exp\left[N\int \tilde{\sigma}_i(t)\left(\partial_t\sigma_i(t) + \mu(t)\sigma_i(t) - Q'(\langle \pmb \sigma(t), \pmb \sigma^*\rangle^{p-1})\sigma^*_i\right)dt\right] \times
		\\
		& \times \mathbb{E}_\eta\left\{\prod_i \exp\left[-\int \tilde{\sigma}_i(t)\left(-\frac{\sqrt{N(p-1)!}}{\Delta_p}\sum_{i_1<\dots<i_p} \eta_{i\,i_1\dots i_{p-1}} \sigma_{i_1}\dots\sigma_{i_{p-1}} \right)dt\right]\right\} \times
		\\
		& \times \mathbb{E}_\eta\left\{\prod_i \exp\left[-\int \tilde{\sigma}_i(t)\left(- \frac{\sqrt{N}}{\Delta_2}\sum_{j} \eta_{ij} \sigma_j \right)dt\right]\right\}\,.
	\end{split}
\end{equation}
In averaging over the $\eta$ we need to be careful in grouping all the permutations of $i$ with $i_1,\dots,i_{p-1}$. For instance the exponent of the term in $p$ is gives by
\begin{equation*}
	\begin{split}
		& -\frac{\sqrt{N(p-1)!}}{\Delta_p} \!\!\!\sum_{i_1<\dots<i_p} \int\eta_{i_1\dots i_p}\left(\tilde{\sigma}_{i_1}(t)\sigma_{i_2}(t)\dots \sigma_{i_p}(t) + \dots + \sigma_{i_1}(t)\sigma_{i_2}(t)\dots \tilde{\sigma}_{i_p}(t) \right)dt
		\\
		&\, = \frac{N(p-1)!}{2\Delta_p} \!\!\!\sum_{i_1<\dots<i_p}\!\! \iint\! \left(\tilde{\sigma}_{i_1}(t)\sigma_{i_2}(t)\dots \sigma_{i_p}(t) \!+\! \text{perm.}\right) \! \left(\tilde{\sigma}_{i_1}(t')\sigma_{i_2}(t')\dots \sigma_{i_p}(t') \!+\! \text{perm.}\right)dtdt'
		\\
		&\, = \frac{N}{2\Delta_p}\!\! \iint\! \left(\langle\tilde{\pmb\sigma}(t),\tilde{\pmb\sigma}(t')\rangle \langle \pmb\sigma(t), \pmb\sigma(t) \rangle^{p-1} \!\!+\! (p-1)\langle\tilde{\pmb\sigma}(t),\pmb\sigma(t')\rangle \langle \pmb\sigma(t),\tilde{\pmb\sigma}(t')\rangle \langle \pmb\sigma(t), \pmb\sigma(t) \rangle^{p-2} \right)dtdt'.
	\end{split}
\end{equation*}
This gives an action $\mathcal{S}[\pmb\sigma,\tilde{\pmb\sigma}]$ defined by
\begin{equation}\label{eq:averaged_generating_functional}
	\begin{split}
		1 & = \overline{\mathcal{Z}} =  \int \mathcal{D}[\pmb\sigma,\tilde{\pmb\sigma}] e^{\mathcal{S}[\pmb\sigma,\tilde{\pmb\sigma}]} =
		\\
		& = \int \mathcal{D}\left[\pmb\sigma,\tilde{\pmb\sigma}\right] \prod_{i}\exp\left[N\int \tilde{\sigma}_i(t)\left(\partial_t\sigma_i(t) + \mu(t)\sigma_i(t) - Q'(\langle \pmb \sigma(t), \pmb \sigma^*\rangle^{p-1})\sigma^*_i \right)dt\right] \times
		\\
		& \times \exp\left[ \frac{N}{2\Delta_p}\iint \langle\tilde{\pmb\sigma}(t),\tilde{\pmb\sigma}(t')\rangle \langle \pmb\sigma(t), \pmb\sigma(t) \rangle^{p-1} dtdt'\right] \times
		\\
		& \times \exp\left[ \frac{N}{2\Delta_p}\iint (p-1)\langle\tilde{\pmb\sigma}(t),\pmb\sigma(t')\rangle \langle \pmb\sigma(t),\tilde{\pmb\sigma}(t')\rangle \langle \pmb\sigma(t), \pmb\sigma(t) \rangle^{p-2}dtdt'\right] \times
		\\
		& \times \exp\left[ \frac{N}{2\Delta_2}\iint \left(\langle\tilde{\pmb\sigma}(t),\tilde{\pmb\sigma}(t')\rangle \langle \pmb\sigma(t), \pmb\sigma(t) \rangle + \langle\tilde{\pmb\sigma}(t),\pmb\sigma(t')\rangle \langle \pmb\sigma(t),\tilde{\pmb\sigma}(t')\rangle \right)dtdt'\right]\,.
	\end{split}
\end{equation}

A simple way to proceed once evaluated the action was proposed in \cite{barrat1997p} and consists in taking the expectation with respect to the path distribution and exploiting simple identities together with integration by part:
\begin{equation}
	\begin{split}
		0 & = - \left\langle \frac{\delta \sigma_i(t')}{\delta \tilde{\sigma}_i(t)}\right\rangle_{\mathcal{S}} = \left\langle \sigma_i(t') \frac{\delta \mathcal{S}}{\delta \tilde{\sigma}_i(t)}\right\rangle_{\mathcal{S}} =
		\\
		& = N\Bigg\langle \partial_t\sigma_i(t) \sigma_i(t') + \mu(t)\sigma_i(t) \sigma_i(t') - Q'(\langle \pmb \sigma(t), \pmb \sigma^*\rangle^{p-1})\sigma^*_i\sigma_i(t') +
		\\
		& + \frac1{\Delta_p}\int \left[\langle \pmb\sigma(t), \pmb\sigma(t) \rangle^{p-1}\tilde{\sigma}_i(t'') + (p-1)\langle\tilde{\pmb\sigma}(t),\pmb\sigma(t'')\rangle \langle \pmb\sigma(t), \pmb\sigma(t) \rangle^{p-2}\sigma_i(t'') \right]dt'' +
		\\
		& + \frac1{\Delta_2}\int \left[\langle \pmb\sigma(t), \pmb\sigma(t) \rangle\tilde{\sigma}_i(t'') + \langle\tilde{\pmb\sigma}(t),\pmb\sigma(t'')\rangle \sigma_i(t'') \right]dt'' \Bigg\rangle_{\mathcal{S}} \,.
	\end{split}
\end{equation}
Finally, summing over the index $i$ and dividing by $N$ we recover Eq.~\eqref{eq:CHSCK_C}.
The remaining CHSCK Eqs.~(\ref{eq:CHSCK_R}-\ref{eq:CHSCK_Cbar}) follow analogously from:
\begin{align}
	 & \delta(t-t') = \sum_i \left\langle \frac{\delta \tilde{\sigma}_i(t')}{\delta \tilde{\sigma}_i(t)} \right\rangle_{\mathcal{S}} = -\sum_i \left\langle \tilde{\sigma}_i(t') \frac{\delta S}{\delta \tilde{\sigma}_i(t)} \right\rangle_{\mathcal{S}} \,;
	\\
	 & 0 = - \left\langle \frac{\delta \sigma_i^*}{\delta \tilde{\sigma}_i(t)}\right\rangle_{\mathcal{S}} = \left\langle \sigma_i^* \frac{\delta \mathcal{S}}{\delta \tilde{\sigma}_i(t)}\right\rangle_{\mathcal{S}}\,.
\end{align}
and Eq.~\eqref{eq:CHSCK_mu} comes from imposing the spherical constrain, $C(t,t)=1\quad\forall t$, on Eq.~\eqref{eq:CHSCK_C}.

In the following we are going to perform the analysis proposed by \cite{CK94SK} in the present model. We need to consider Langevin dynamics instead of gradient-flow dynamics
\begin{equation}\label{eq:Langevin_dynamics}
	\dot\sigma_i(t) = -\mu(t)\sigma(t) - \frac{\partial \loss(\pmb\sigma(t) | \pmb T, \pmb Y)}{\partial \sigma_i(t)} + \frac1{\sqrt N}\eta_i^{(L)}(t)\,,
\end{equation}
where the last term represents the Langevin noise, which is white Gaussian noise with moments: $\mathbb{E}_{L}[\eta_i^{(L)}(t)] = 0$ and $\mathbb{E}_{L}[\eta_i^{(L)}(t)\eta_j^{(L)}(t')] = 2T\delta_{ij}\delta(t-t')$ with $T$ that has the physical meaning of temperature. The CHSCK equations are slightly modified,
\begin{align}
	\begin{split}
		&\frac{\partial}{\partial t} C(t,t') = T R(t',t) - \mu(t)\,C(t,t')+
		Q'(m(t)) m(t') +
		\\
		&\quad+ \int_0^t R(t,t'')Q''(C(t,t''))C(t',t'') dt''
		+ \int_0^{t'} R(t',t'')Q'(C(t,t'')) dt''\,,
	\end{split}\label{eq:CHSCK_C_T}
	\\
	\begin{split}
		&\frac{\partial}{\partial t} R(t,t')= - \mu(t)\,R(t,t') +\int_{t'}^tR(t,t'')Q''(C(t,t''))R(t'',t') dt''\,,
	\end{split}\label{eq:CHSCK_R_T}
	\\
	\begin{split}
		&\frac{\partial}{\partial t}  m(t) = -\mu(t)\,m(t)+Q'(m(t)) + \int_{0}^t R(t,t'')m(t'') Q''(C(t,t'')) dt''\,,
	\end{split}\label{eq:CHSCK_Cbar_T}
	\\
	\begin{split}
		& \mu(t) = T + Q'(m(t))m(t) + \int_{0}^t R(t,t'')\left[Q'(C(t,t'')) + Q''(C(t,t''))C(t,t'')\right] dt''\,.
	\end{split}\label{eq:CHSCK_mu_T}
\end{align}

\subsection{CHSCK Equations Separation of Time-Scales}\label{sec:CHSCK_CK_analysis}

\begin{figure}[t!]
	\centering
	\includegraphics[width=0.5\linewidth]{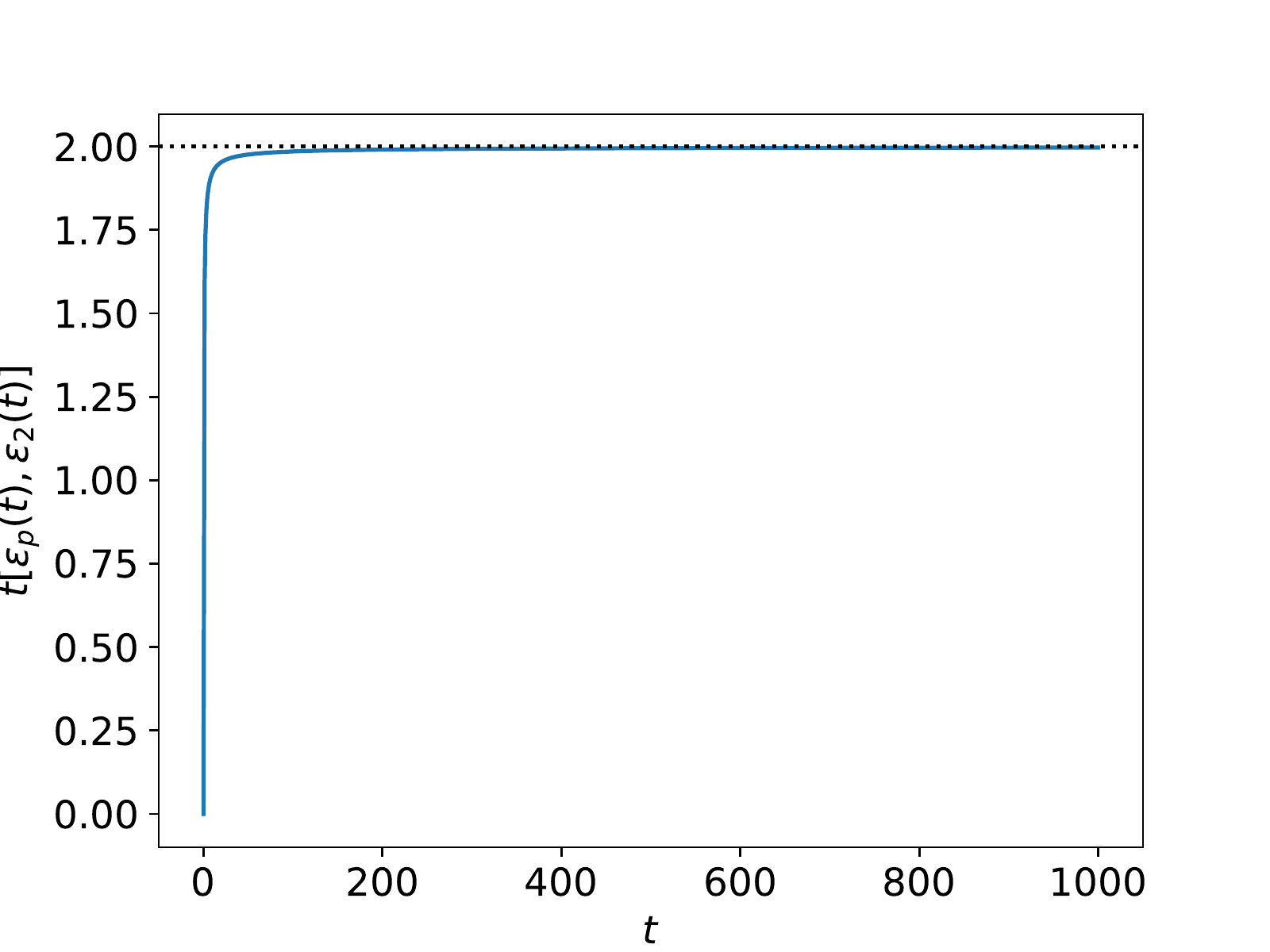}
	\caption{$t = -\left(p\epsilon_p(t)+2\epsilon_2(t)\right)/\sqrt{Q''(1)}$ for $p=3$, $\Delta_p = 1.0$ and $1/\Delta_2 = 1.9$ evaluated numerically from the CHSCK equations.}
	\label{fig:t_ag}
\end{figure}

\begin{figure}[t!]
	\centering
	\includegraphics[width=0.5\linewidth]{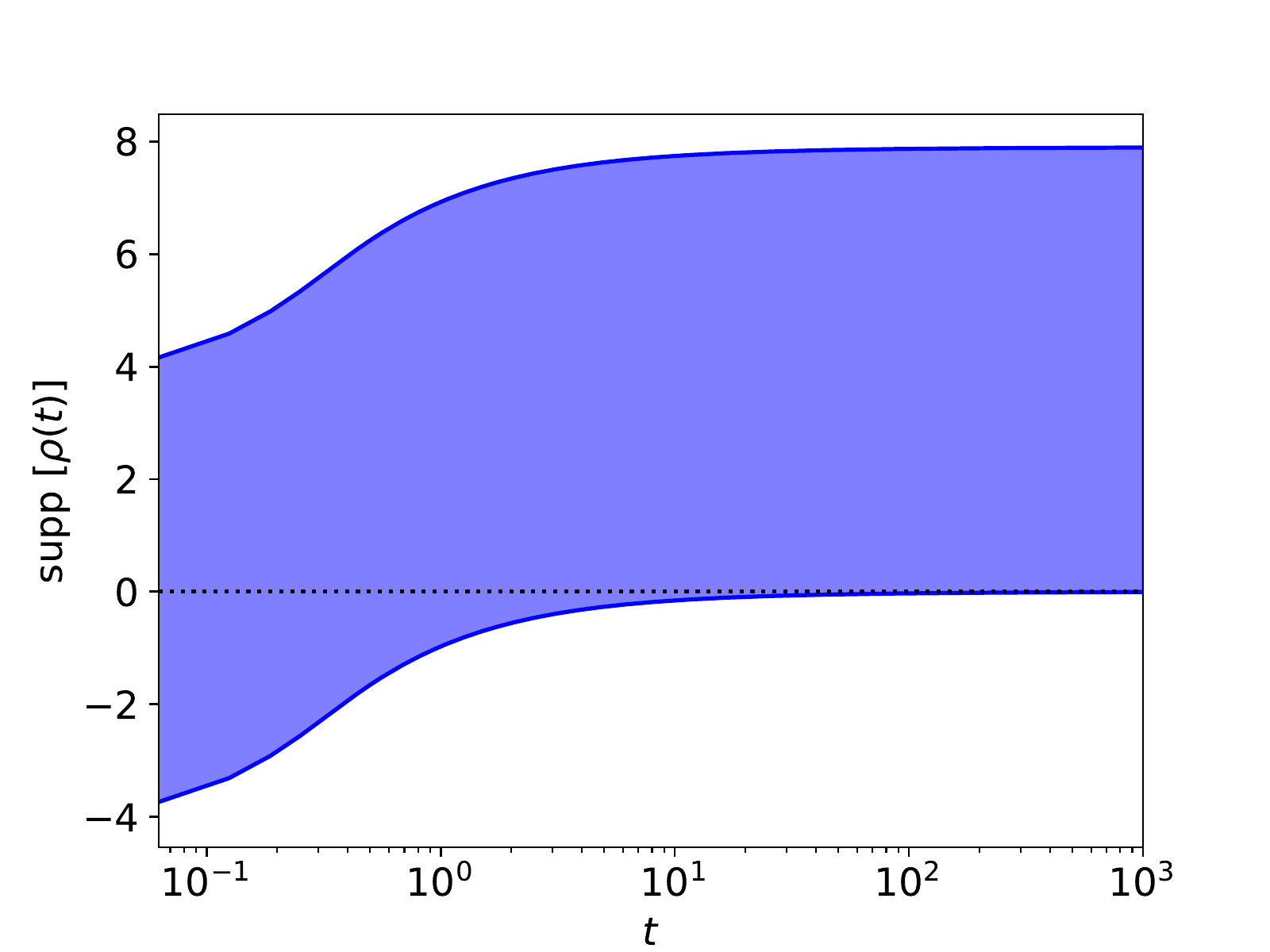}
	\caption{Support of the density of eigenvalues of the Hessian along the dynamics for $p=3$, $\Delta_p = 1.0$ and $1/\Delta_2 = 1.9$. This figure illustrates the GF tends to the marginally stable minima for low signal-to-noise ratio.}
	\label{fig:hess_ag}
\end{figure}

\begin{figure}[t!]
	\centering
	\includegraphics[width=0.5\linewidth]{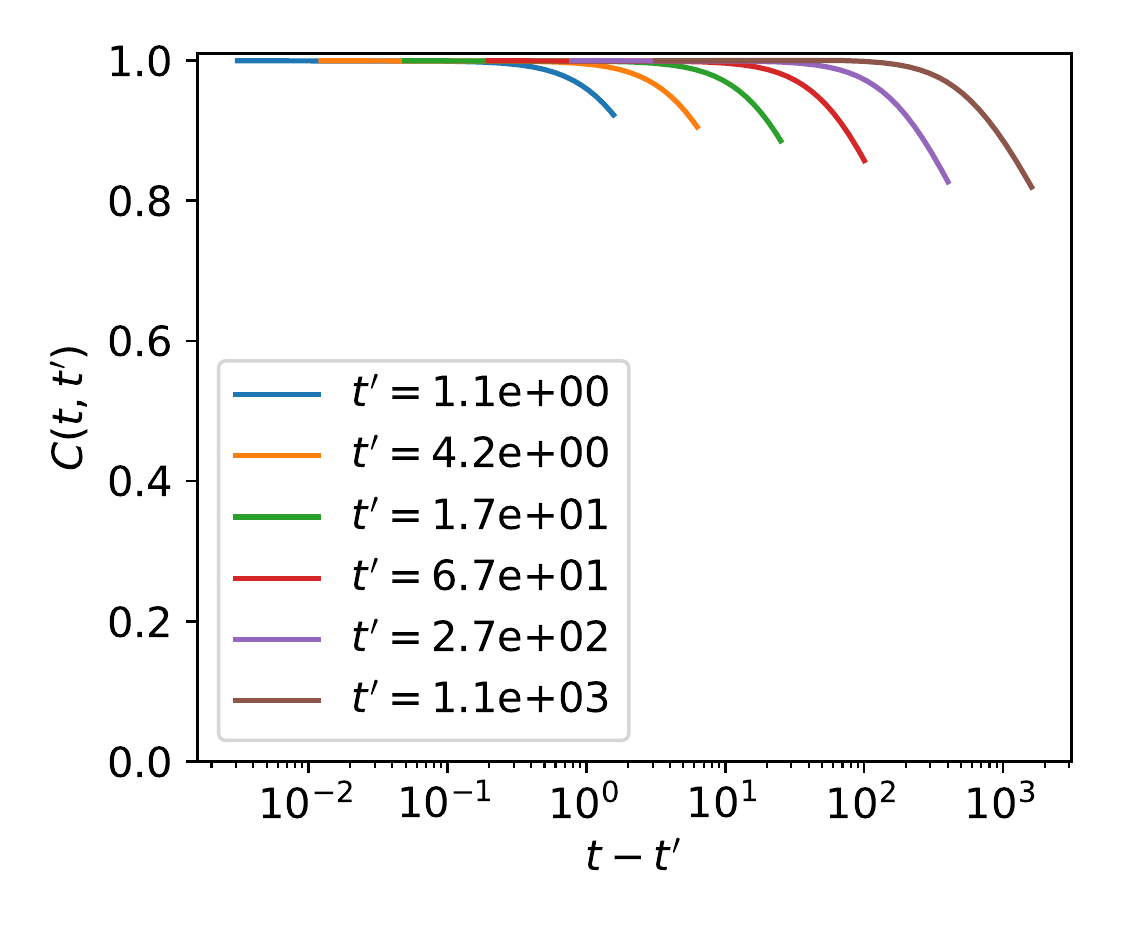}
	\caption{The correlation function $C(t,t')$ with $p=3$, $\Delta_p = 1.0$ and $1/\Delta_2 = 1.5$ evaluated numerically from the CHSCK equations. The correlation is plotted as difference of the two times showing the as $t-t'\ll t, t'$ it remains close to $1$. This shows as well that in this regime the correlation function shows time translational invariance.}
	\label{fig:correlation_fdt}
\end{figure}

The theory of glassy dynamics \cite{cugliandolo1993analytical} is quite involved. 
We have therefore decided to show first some numerical results that directly confirm assumptions made in the main text, and then show in full glory that these assumptions can be obtained analytically from the theory. 

\subsubsection{Numerical tests}
The first property that we wish to test is that for low signal-to-noise ratio GF is trapped in minima that are marginally stable. This can be checked computing from the CHSCK equations the evolution of $t = -\left(p\epsilon_p(t)+2\epsilon_2(t)\right)/\sqrt{Q''(1)}$, the terms $\epsilon_p(t)$ and $\epsilon_2(t)$ can be expressed in terms of $C,R,m$ as shown in Rmk.~\ref{remark:energy} of section~\ref{sec:parameters}. 
In Fig.~\ref{fig:t_ag} we show that $t = -\left(p\epsilon_p(t)+2\epsilon_2(t)\right)/\sqrt{Q''(1)}$ ($\epsilon_p(t)$ starts from zero at initial time and then converges to 
two. Thus the minima to which GF tends to at long times and for small signal-to-noise ratio are indeed the {\it marginally stable ones} characterized by a spectrum of the 
Hessian whose left edge touches zero. Actually, transferring the results obtained in the context of spin-glasses to our case \cite{kurchan1996phase}, we know that as long as $m$ remains zero, i.e for small  signal-to-noise ratio, the spectrum of the Hessian along the dynamics is a Wigner semi-circle  with support $[\sqrt{Q''(1)}(-2+t),\sqrt{Q''(1)}(2+t)]$. We show the evolution of the support as a function of time in Fig.~\ref{fig:hess_ag}. This is another illustration of the fact that minima trapping GF are marginally stable. 

The other point we wish to test is the assumption (i) made in the main text on $C(t,t')$, which we repeat here for convenience: $C(t,t')=1$ when $t-t'$ finite; $C(t,t')$ becomes less than one when $t-t'$ diverges with $t$ and $t'$.
We show in Fig.~\ref{fig:correlation_fdt} the correlation function $C(t,t')$ as a function of $t-t'$ (in log-scale) for several values of $t'$. This is a good illustration of the "aging ansatz" defined in the main text. 

Let us stress that these numerical tests where already done in the past on similar spin-glass models. We show them here in order to make the paper self-contained 
and so that the reader does not have to go back to physics literature. In the same vein, in the next sections we show the full theoretical analysis of the dynamical 
equations, which closely follows the theory of glassy dynamics developed in physics \cite{cugliandolo1993analytical}.  

\subsubsection{General aging ansatz}

We study the behaviours of the dynamics at large times in the two-time regimes introduced in the main text (now generalized at finite temperature).
\begin{enumerate}
	\item $t, t' \gg 1$ with $\frac{t-t'}t\rightarrow0$, see Fig.~\ref{fig:correlation_fdt}. In this regime we have two important aspects: the two-times function depends only on the difference of the two times, $\tau = t-t'$, and we say that they respect time-translational invariance. Under this observation we redefine the two functions as $C(t,t') \rightarrow \Cfdt(\tau)\equiv C(t-t',0)$ and $R(t,t') \rightarrow \Rfdt(\tau)\equiv R(t-t',0)$.  The second important aspect is the validity of the fluctuation-dissipation theorem (FDT) that links correlation and response function by the relation $\Rfdt(\tau) = -\frac1T \frac{d\Cfdt}{d\tau}(\tau)$ for $\tau$ positive.
	\item $t, t' \gg 1$ with $\frac{t-t'}t=O(1)$. In this regime the relevant variable to consider is $\lambda=\frac{t'}{t}$. In reason of the "weak-long term memory" property it is useful to redefine rescale the response function and define $\mathcal{R}(\lambda)=tR(t,t')$. It is also convenient to consider the function $q\mathcal{C}(\lambda)=C(t,t')$ with $q = \lim_{\tau\rightarrow\infty}\Cfdt(\tau)$. Finally, in this regime a generalised version of the fluctuation-dissipation theorem holds $\mathcal{R}(\lambda)=\frac1T x q \frac{d\mathcal{C}(\lambda)}{d\lambda}$ where $x$ is called {\it violation parameter}.
\end{enumerate}
Under the (generalized) FDT the equations for correlation and response function that we obtain in the two-time regime collapse into a single equation. In the first regime we analyse only the correlation because of this link, while in the second regime we consider the two equations separately since we need to determine the violation parameter $x$.

In the analysis that follows we use massively the hypothesis of the two regimes to split the integrals and analyse them separately. We start analysing the single time equation for the Lagrange multiplier $\mu(t)$, then we proceed with the two-times function concentrating first on the time-translational invariant part and then on the aging part.

\subsubsection{Langrange multiplier in the large time limit.} Starting from Eq.~\eqref{eq:CHSCK_mu_T}, we introduce the symbol $\clubsuit_p$ to isolate the two contribution of matrix and tensor to the integral. As the time tends to infinity $m$ and $\mu$ tend to their asymptotic value, respectively $m_\infty$ and $\mu_\infty$, Eq.~\eqref{eq:CHSCK_mu_T} tends to
\begin{equation*}
	\mu_\infty = T +Q'(\Cmag_\infty)\Cmag_\infty + p\clubsuit_{p}+ 2\clubsuit_{p=2}\,.
\end{equation*}
We can now use the idea of the separation in two-time regimes. Call $Q_p(x) = x^p/(p\Delta_p)$ the part related to $p$ in $Q(x)$,
\begin{align*}
	\clubsuit_p = \int_0^t & Q_p'(C(t,t''))R(t,t'') dt'' = \int_\text{FDT} + \int_\text{aging} =                                                                                \\
	                       & = -\int_t^0 Q_p'(C(t,t-\tilde{t}))R(t,t-\tilde{t}) d\tilde{t} + \int_0^1 \mathcal{R}(\lambda)Q_p'(q\mathcal{C}(\lambda)) d\lambda =                \\
	                       & = +\int_0^\infty Q_p'(\Cfdt(\tilde{t}))\Rfdt(\tilde{t}) d\tilde{t} + \int_0^1 \mathcal{R}(\lambda) Q_p'(q\mathcal{C}(\lambda)) d\lambda =          \\
	                       & = -\int_0^\infty \frac1T\frac{d}{d\tilde{t}}Q_p(\Cfdt(\tilde{t})) d\tilde{t} + \int_0^1 \mathcal{R}(\lambda)Q_p'(q\mathcal{C}(\lambda)) d\lambda = \\
	                       & = \frac{1-q^p}{T\Delta_p}
	+ \int_0^1 \mathcal{R}(\lambda)Q_p'(q\mathcal{C}(\lambda)) d\lambda\,
\end{align*}

the resulting equation is
\begin{equation}
	\begin{split} \label{eq:mu_infinity}
		\mu_\infty &=T +Q'(\Cmag_\infty)\Cmag_\infty + \frac1{T}\left[Q'(1)-q\,Q'(q)\right]+\\
		&+p\int_0^1 \mathcal{R}(\lambda)Q_p'(q\mathcal{C}(\lambda)) d\lambda+2\int_0^1 \mathcal{R}(\lambda)Q_2'(q\mathcal{C}(\lambda)) d\lambda\,.
	\end{split}
\end{equation}

\subsubsection{Regime 1: FDT.} We apply the same scheme of separating the times scale and applying FDT to the correlation function. All over the analysis we isolate terms in the equations using the symbols $\clubsuit$ and $\spadesuit$. Eq.~\eqref{eq:CHSCK_C_T} in the large time is
\begin{equation}\label{eq:Cfdt_sketch}
	(\partial_\tau+\mu_\infty)\Cfdt(\tau) = Q'(\Cmag_\infty)\Cmag_\infty + \spadesuit + \clubsuit\,,
\end{equation}
with:
\begin{align*}
	\clubsuit & = \int_0^{t'} Q'(C(t,t''))R(t',t'') dt'' = \int_\text{FDT} + \int_\text{aging} =                                                                                    \\
	          & = -\int_{t'}^0 Q'(C(t,t'-\tilde{t}))R(t',t'-\tilde{t}) d\tilde{t} + \int_0^1 Q'(q\mathcal{C}(\lambda))\mathcal{R}(\lambda) d\lambda =                               \\
	          & = -\frac1T\int_0^\infty Q'(\Cfdt(\tau+\tilde{t})) \frac{d}{d\tilde{t}}\Cfdt(\tilde{t}) d\tilde{t} + \int_0^1 Q'(q\mathcal{C}(\lambda))\mathcal{R}(\lambda) d\lambda
\end{align*}
and
\begin{align*}
	\spadesuit & = \int_0^{t} Q''(C(t,t''))R(t,t'')C(t',t'') dt'' = \int_{t'}^t + \int_0^{t'} =
	\int_{t'}^t + \int_\text{FDT} + \int_\text{aging} =                                                                                                                                                      \\
	           & = \frac1T\left[Q'(1)\Cfdt(\tau) - Q'(q)q\right] -\frac1T\int_0^\tau Q'(\Cfdt(\tau-\tilde{t}))\frac{d}{d\tilde{t}}\Cfdt(\tilde{t}) d\tilde{t}                                                \\
	           & +\frac1T\int_0^\infty Q'(\Cfdt(\tau+\tilde{t}))\frac{d}{d\tilde{t}}\Cfdt(\tilde{t}) d\tilde{t} + \int_0^1 \mathcal{R}(\lambda)Q''(q\mathcal{C}(\lambda))\,q\mathcal{C}(\lambda) d\lambda\,.
\end{align*}


Substituting these expressions and using Eq.~\eqref{eq:mu_infinity} in Eq.~\eqref{eq:Cfdt_sketch} we have the first equation, which characterises the first regime
\begin{equation}\label{eq:CHSCK_C_fdt}
	\partial_\tau \Cfdt(\tau) + \Big(\frac1T Q'(1) - \mu_\infty\Big) \left[ 1-\Cfdt(\tau) \right] + T = - \frac1T\int_0^\tau Q'(\Cfdt(\tau-\tau''))\frac{d}{d\tau''}\Cfdt(\tau'') d\tau''\,.
\end{equation}
An important limit that is used later on in the computation is when $\tau\rightarrow\infty$ and the variations of $\Cfdt$ becomes irrelevant. This gives:
\begin{equation} \label{eq:mu_infinity_2}
	\mu_\infty = \frac{T}{1-q}+\frac{Q'(1)-Q'(q)}T\,.
\end{equation}

\subsubsection{Regime 2: aging.} Starting from the evolution of the response function \eqref{eq:CHSCK_R_T}, in this regime the time derivative is negligible.

\begin{equation*}
	0 = -\mu_\infty \frac{\mathcal{R}(\lambda)}t+ \clubsuit
\end{equation*}

with $\clubsuit$ that can be separated into three terms $\clubsuit^{(1)}$, $\clubsuit^{(2)}$ and $\clubsuit^{(3)}$

\begin{equation*}
	\clubsuit = \int_{t'}^t R(t,t'')Q''(C(t,t''))R(t'',t') dt'' = \int_{t''\lesssim t} + \int_{t''\gtrsim t'} + \int_\text{aging} = \clubsuit^{(1)}+\clubsuit^{(2)}+\clubsuit^{(3)}\,.
\end{equation*}
In the first two integrals we can apply FDT respectively for $t''$ close to $t$ and for $t''$ close to $t'$:
\begin{equation*}
	\begin{split}
		\clubsuit^{(1)} &= \int_0^\infty R(t,t-\tilde{t})Q''(C(t,t-\tilde{t}))\frac{\mathcal{R}(\lambda)}t d\tilde{t} = -\frac{\mathcal{R}(\lambda)}t\frac1T\int_0^\infty \frac{d}{d\tilde{t}}Q'(\Cfdt(\tilde{t})) d\tilde{t} = \\
		& = \frac1T \frac{\mathcal{R}(\lambda)}t \left(Q'(1) -Q'(q)\right)\,,
	\end{split}
\end{equation*}
\begin{equation*}
	\begin{split}
		\clubsuit^{(2)} &= \int_0^\infty \frac1t\mathcal{R}(\lambda)Q''(q\mathcal{C}(\lambda))R(t'+\tilde{t},t') d\tilde{t} = -\frac{\mathcal{R}(\lambda)}t\,Q''(q\mathcal{C}(\lambda))\frac1T\int_0^\infty \frac{d}{d\tilde{t}}\Cfdt(\tilde{t}) d\tilde{t} = \\
		&= \frac1T\frac{\mathcal{R}(\lambda)}t\,Q''(q\mathcal{C}(\lambda))(1-q)\,.
	\end{split}
\end{equation*}
The last terms displays aging:
\begin{equation*}
	\clubsuit^{(3)} = \int_{t'}^t \frac{\mathcal{R}(\frac{t''}t)}t\frac{\mathcal{R}(\frac{t'}{t''})}{t''}Q''\left(q\mathcal{C}\left(\frac{t''}t\right)\right) dt'' = \frac1t \int_{\lambda}^1 \frac{\mathcal{R}(\lambda'')}{\lambda''}\mathcal{R}\left(\frac{\lambda}{\lambda''}\right)Q''\left(q\mathcal{C}(\lambda'')\right) d\lambda''\,.
\end{equation*}
Combining these pieces together and using \eqref{eq:mu_infinity_2} we obtain an expression for the aging function of the response:
\begin{equation}\label{eq:CHSCK_R_aging}
	0 = \left[ -\frac{T}{1-q}+ \frac{Q''(q\mathcal{C}(\lambda))(1-q)}T \right] \mathcal{R}(\lambda) + \int_{\lambda}^1 \frac{\mathcal{R}(\lambda'')}{\lambda''}\mathcal{R}\left(\frac{\lambda}{\lambda''}\right)Q''\left(q\mathcal{C}(\lambda'')\right) d\lambda'' \, .
\end{equation}
Following the same steps in Eq.~\eqref{eq:CHSCK_C_T}
and using again \eqref{eq:mu_infinity_2} we obtain the expression for the correlation:
\begin{equation}\label{eq:CHSCK_C_aging}
	\begin{split}
		0 &= - \left[ \frac{T}{1-q} + \frac{Q'(q\mathcal{C}(\lambda))(1-q)}{q\,\mathcal{C}(\lambda)\,T} \right] q\, \mathcal{C}(\lambda)+
		Q'(m_\infty) m_\infty +
		\\
		& +\int_0^\lambda Q'(q\mathcal{C}(\lambda''))\mathcal{R}\left(\frac{\lambda''}\lambda\right) \frac{d\lambda''}{\lambda} + q\int_0^1 \mathcal{R}(\lambda'') Q''(q\mathcal{C}(\lambda'')) \mathcal{C}\left[\left(\frac{\lambda''}\lambda\right)^{\text{sign}(\lambda-\lambda'')}\right] d\lambda''
	\end{split} \, .
\end{equation}

\begin{remark}[generalized-FDT]
	In the derivation we never used generalized-FDT ansatz, $\mathcal{R}(\lambda)=\frac1T x q \frac{d\mathcal{C}(\lambda)}{d\lambda}$. A posteriori we can observe its validity as Eq.~\eqref{eq:CHSCK_R_aging} and Eq.~\eqref{eq:CHSCK_C_aging} collapse to a single equation as Eq.~\eqref{eq:CHSCK_C_aging} is derived by $\lambda$ and generalized-FDT is used.
\end{remark}

\subsubsection{Characterisation of the unknown parameters.} 
\label{sec:parameters}
In order to fully characterized the FDT Eq.~\eqref{eq:CHSCK_C_fdt} and the aging Eqs.~(\ref{eq:CHSCK_R_aging}-\ref{eq:CHSCK_C_aging}), we need to determine the parameters $\Cmag_\infty$, $\mu_\infty$, $q$, the FDT violation index $x$ and $q_0 = q\mathcal{C}(0)$. We do not determine all of them, we consider only the few that are used in the analysis, but for sake of completeness we say how the five equations can be determined: Eq.~\eqref{eq:CHSCK_Cbar_T} taking $t\rightarrow\infty$, Eq.~\eqref{eq:mu_infinity} plugging the generalized FDT ansatz, Eq.~\eqref{eq:CHSCK_C_fdt} in the large $\tau$ limit, Eq.~\eqref{eq:CHSCK_R_aging} in the limit $\lambda\rightarrow1$, Eq.~\eqref{eq:CHSCK_C_aging} in the limit $\lambda\rightarrow0$.

In particular the $\lim\lambda\rightarrow1$ of Eq.~\eqref{eq:CHSCK_R_aging} gives
\begin{equation}\label{eq:auxiliary_2}
	\frac{T^2}{(1-q)^2} = 
	Q''(q)\,.
\end{equation}

From Eq.~\eqref{eq:CHSCK_C_aging}, in the limit $\lambda \rightarrow 1$ and $\lambda \rightarrow 0$ we obtain
\begin{align}
	 & 0 = \left[\frac{Tq}{1-q} - \frac{Q'(q)(1-q)}T\right] + Q'(m_\infty)m_\infty + \frac{x}T [q\,Q'(q) - q_0\,Q'(q_0)]\,,
	\\
	 & 0 = \left[\frac{Tq_0}{1-q} - \frac{Q'(q_0)(1-q)}T\right] + Q'(m_\infty)m_\infty + q_0 \frac{x}T [Q'(q) -\,Q'(q_0)]\,.
\end{align}
In the regime where the system does not find a good overlap with the signal thus $m_\infty = 0$, the second equation gives the solution $q_0 = 0$. As $T$ tends to $0$ (and $q$ tends to $1$)
\begin{equation}\label{eq:parisi_parameter/T}
	\frac{x}T = \frac1{q\,Q'(q)}\left[\frac{T}{1-q} - \frac{Q'(1)(1-q)}T\right] = \frac1{q\,Q'(q)}\left[\sqrt{Q''(q)} - \frac{Q'(1)}{\sqrt{Q''(q)}}\right]\,.
\end{equation}

\begin{remark}[$\overline{R}$]
	In the large time limit, and using FDT, we have
	\begin{equation}
		\overline{R} = \int_0^\infty \Rfdt(\tau'') d\tau'' = -\frac1T\int_0^\infty \Cfdt'(\tau'')d\tau'' = \frac{1-q}T\,,
	\end{equation}
	using Eq.~\eqref{eq:auxiliary_2} as $T\rightarrow0$ we the result reported in the main text
	\begin{equation}
		\overline{R} = \frac1{\sqrt{Q''(1)}}\,.
	\end{equation}
\end{remark}

\begin{remark}[marginal states]
	Combining Eq.~\eqref{eq:auxiliary_2} with Eq.~\eqref{eq:mu_infinity_2}, we obtain:
	\begin{equation}
		\mu_\infty = \sqrt{Q''(q)} + \frac1T\left[Q'(1) - Q'(q)\right]
	\end{equation}
	expanding $q\lessapprox1$, and using again Eq.~\eqref{eq:auxiliary_2},
	\begin{equation}
		\mu_\infty = 2\sqrt{Q''(1)}\,.
	\end{equation}
	As we explained in the main text, the distribution of the Hessian is associated to a semicircle of radius $2\sqrt{Q''(1)}$ and centred in $\mu$. This equation tells that asymptotically, if aging does not stops -- as it happens if it jumps to the solution -- the systems tends to the marginal states. We have shown a numerical confirmation of this property in Fig.~\ref{fig:t_ag}.
\end{remark}

\begin{remark}[threshold loss] \label{remark:energy}
    As we show in the main text the Lagrange multiplier $\mu$ depends on the two losses as $\mu = -p\epsilon_p - 2\epsilon_2$ (or $\mu = T - p\epsilon_p - 2\epsilon_2$ for arbitrary temperature). Observing that the equation holds for any $\Delta_p$ and $\Delta_2$, in particular when they tend to infinity and therefore their contribution to the total loss becomes irrelevant, it follows from Eq.~\eqref{eq:CHSCK_mu} (respectively Eq.~\eqref{eq:CHSCK_mu_T}),
	\begin{equation}\label{eq:CHSCK_energy_p}
		\epsilon_p(t) = -\frac1p \left[ Q_p'(m(t))m(t) + \int_{0}^t R(t,t'')\left[Q_p'(C(t,t'')) + Q_p''(C(t,t''))C(t,t'')\right] dt''\right]\,
	\end{equation}
	and analogously $\epsilon_2(t)$. We then write the expression for the total loss
	\begin{equation}\label{eq:CHSCK_energy}
	    \begin{split}
		    \epsilon(t) &= -\frac1p Q_p'(m(t))m(t) -\frac12 \left[ Q_2'(m(t))m(t)\right] + 
		    \\
		    &+ \int_{0}^t R(t,t'')\left[Q'(C(t,t'')) + Q''(C(t,t''))C(t,t'')\right] dt''
		    \,.
		\end{split}
	\end{equation}
    
\end{remark}

From the equation we established above and using the aging ansatz, one can obtain the asymptotic value of the loss for low signal-to-noise ratio, i.e. the loss of the minima trapping the dynamics.

\begin{remark}[threshold energy] 
    The large time limit of Eq.~\eqref{eq:CHSCK_energy} gives two threshold states. Applying the same scheme used in Eq.~\eqref{eq:mu_infinity}, \textit{i.e.} integrating Eq.~\eqref{eq:CHSCK_energy_p} for $t, t' \gg 1$ considering the two time-regimes gives
	\begin{equation}
		\begin{split}
			\epsilon_{p,\text{\rm th}}^\text{\rm dyn} &= -\frac1p \left[ Q_p'(\Cmag_\infty)\Cmag_\infty + \frac1{T}\left[Q_p'(1)-q\,Q_p'(q)\right]
			+p\int_0^1 \mathcal{R}(\lambda)Q_p'(q\mathcal{C}(\lambda)) d\lambda\right] \,.
		\end{split}
	\end{equation}
	Applying the generalized fluctuation dissipation ansatz and Eq.~\eqref{eq:parisi_parameter/T} in the integral, and finally taking $T\rightarrow0$ ($q\rightarrow1$)
	\begin{equation}\label{eq:threshold_energy_dyn}
		\epsilon_{p,\text{\rm th}}^\text{\rm dyn} = - \frac1p\frac{Q_p'(1) + Q_p''(1)}{\sqrt{Q''(1)}} - \frac{x}T Q_p(1)\,.
	\end{equation}
	The threshold energy will be given by the some of two contributions, giving
	\begin{equation}
		\epsilon_\text{\rm th}^\text{\rm dyn} = - \frac{Q'(1)}{\sqrt{Q''(1)}} - \frac{Q(1) \left(Q''(1) - Q'(1)\right)}{\sqrt{Q''(1)} Q'(1)}\,.
	\end{equation}
\end{remark}

\begin{remark}[threshold energies equivalence]
	After some manipulation we notice that the threshold energy evaluated using Kac-Rice formula, Eq.~\eqref{eq:threshold_energy_KR}, and the threshold energy evaluated using the dynamical equations, Eq.~\eqref{eq:threshold_energy_dyn}, are equivalent and therefor the two methods are coherent.
\end{remark}



\section{Numerical Simulations of Gradient-Flow}\label{sec:numerical_study}

\begin{figure}[ht!]
	\centering
	\includegraphics[scale=0.5]{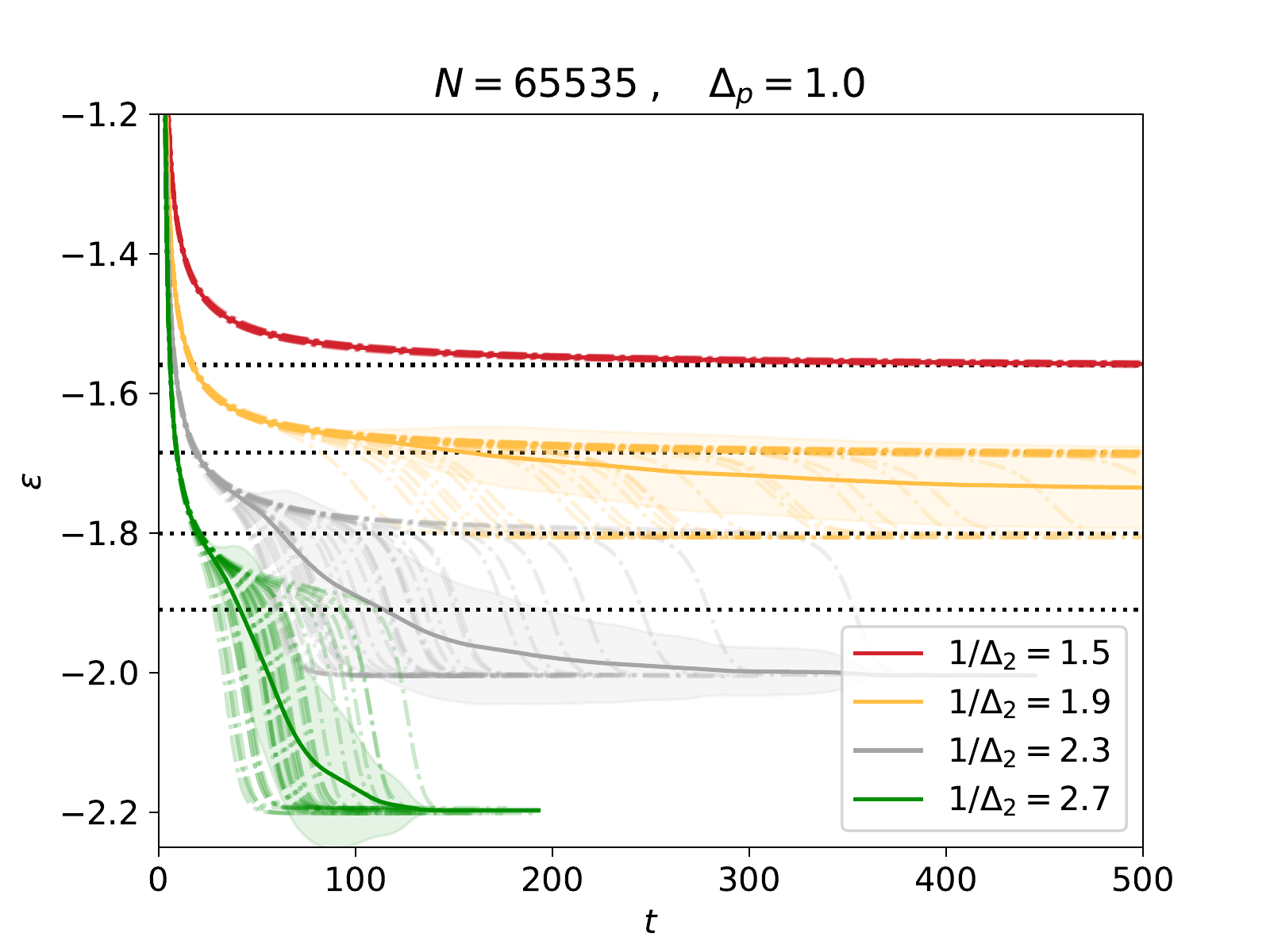}
	\caption{Evolution of the loss in time from numerical simulations realised over 100 instances of disorder and noise, for the spiked matrix-tensor model with $p=3$. The simulations has been done with systems of size $N = 2^{16}-1 = 65535$ with parameters $1/\Delta_2\in\{1.5, 1.9, 2.3, 2.7\}$ and $\Delta_p = 1.0$. 
	}
	\label{fig:numerical_simulation}
\end{figure}

In order to evaluate Gradient flow dynamics we discretized time and evaluated Eq.~\eqref{eq:gradient_flow} numerically using effectively gradient descent
\begin{equation}\label{eq:gradient_descent}
	\sigma_i^{t+1} = -\mu^t \sigma_i^t - \frac{\partial \loss(\pmb\sigma^t | \pmb T, \pmb Y)}{\partial \sigma_i^t}\,.
\end{equation}
In our experiments we run the dynamics on numerous realisations of the problem for different values of the parameters at $p=3$. Given a signal $\pmb\sigma^*\in\mathbb{S}^{N-1}$, the number of computations per interaction scales as $N^3$, which makes the system hard to simulate for large $N$. In order to increase the size of the system, we considered a diluted system, as proposed in  \cite{krzakala2013performance}, instead of the real system, such that the first and the second moment of the loss, in the leading order on $N$. In the original system the (hyper)-graph of interaction is fully connected and counts $N^3/3!$ (hyper)-edges for the tensor and $N^2/2$ edges for the matrix. In the diluted systems we replace the (hyper)-graphs by graphs less connected in particular we take $N^2$ (hyper)-edges for the tensor and $N\sqrt{N}$ edges for the matrix. In systems with spherical variables there is a known problem \cite{semerjian2004stochastic} associated with reducing too much the number of interaction. In general given a tensor of order $p$ if the number of interaction becomes less then $N^{p-1}$ the system tends to favour a finite number of (hyper)-edges and aligns completely with them. The dynamics then converges to a final configuration where $O(p)$ spins have value of order $O(1)$ and the rest is of order $O(1/\sqrt{N})$.
In order to have the same averages for the observables --- such as overlap with the signal and loss --- called $\#(\cdot)$ that counts the number of interactions, we multiplied the variances of the noise by $N^{3/2}/(3!\#(\pmb T))$ and $N/(2 \#(\pmb Y))$ respectively the tensor noise and the matrix noise.

Using this observation in the code we obtain a simple algorithm that given a $dt$ approximate gradient flow by computing a gradient descent dynamics, with $dt = 1.0$ in the simulations. This value was chosen observing that in the runs the algorithm always descends in terms of loss and not appreciable difference appeared reducing it further. The code is made available and attached to this paper. 
Using this code we were able to simulate systems of the size $N=2^{16}-1=65535$ and reduce finite size effects. Fig.~\ref{fig:numerical_simulation} shows the average over different initialisation and realisation of the noise for the parameters presented in the paper $\Delta_p = 1.0$ and  $1/\Delta_2 \in\{1.5,\,,1.9,\,,2.3,\,,2.7,\}$. In the figure we use a continuous line surrounded by a shadow to represent mean and standard deviation under a Gaussian hypothesis, individual simulations are represented using dashed-dotted lines. For $p=3$ and $\Delta_p = 1.0$ the critical threshold for gradient flow occurs at $1/\Delta_2^{\rm GF} = 2.0$ and in fact we observe that the green line ($1/\Delta_2=1.9$) shows finite size effects and some simulations find good overlap with the ground truth. To conclude the figure shows a very good agreement with the averaged value evaluated using CHSCK equations, see Fig.~\ref{fig:dynamics_complexity}-b. In particular is evident how all the dynamics tends to the threshold states, whose corresponding losses are drown with horizontal dotted lines, before eventually find the good direction and then the signal.


\begin{thebibliography}{79}%
\makeatletter
\providecommand \@ifxundefined [1]{%
 \@ifx{#1\undefined}
}%
\providecommand \@ifnum [1]{%
 \ifnum #1\expandafter \@firstoftwo
 \else \expandafter \@secondoftwo
 \fi
}%
\providecommand \@ifx [1]{%
 \ifx #1\expandafter \@firstoftwo
 \else \expandafter \@secondoftwo
 \fi
}%
\providecommand \natexlab [1]{#1}%
\providecommand \enquote  [1]{``#1''}%
\providecommand \bibnamefont  [1]{#1}%
\providecommand \bibfnamefont [1]{#1}%
\providecommand \citenamefont [1]{#1}%
\providecommand \href@noop [0]{\@secondoftwo}%
\providecommand \href [0]{\begingroup \@sanitize@url \@href}%
\providecommand \@href[1]{\@@startlink{#1}\@@href}%
\providecommand \@@href[1]{\endgroup#1\@@endlink}%
\providecommand \@sanitize@url [0]{\catcode `\\12\catcode `\$12\catcode
  `\&12\catcode `\#12\catcode `\^12\catcode `\_12\catcode `\%12\relax}%
\providecommand \@@startlink[1]{}%
\providecommand \@@endlink[0]{}%
\providecommand \url  [0]{\begingroup\@sanitize@url \@url }%
\providecommand \@url [1]{\endgroup\@href {#1}{\urlprefix }}%
\providecommand \urlprefix  [0]{URL }%
\providecommand \Eprint [0]{\href }%
\providecommand \doibase [0]{http://dx.doi.org/}%
\providecommand \selectlanguage [0]{\@gobble}%
\providecommand \bibinfo  [0]{\@secondoftwo}%
\providecommand \bibfield  [0]{\@secondoftwo}%
\providecommand \translation [1]{[#1]}%
\providecommand \BibitemOpen [0]{}%
\providecommand \bibitemStop [0]{}%
\providecommand \bibitemNoStop [0]{.\EOS\space}%
\providecommand \EOS [0]{\spacefactor3000\relax}%
\providecommand \BibitemShut  [1]{\csname bibitem#1\endcsname}%
\let\auto@bib@innerbib\@empty

\bibitem{kawaguchi2016deep}
Kenji Kawaguchi.
\newblock Deep learning without poor local minima.
\newblock In {\em Advances in Neural Information Processing Systems}, pages
  586--594, 2016.

\bibitem{soudry2016no}
Daniel Soudry and Yair Carmon.
\newblock No bad local minima: Data independent training error guarantees for
  multilayer neural networks.
\newblock {\em arXiv preprint arXiv:1605.08361}, 2016.

\bibitem{ge2016matrix}
Rong Ge, Jason~D Lee, and Tengyu Ma.
\newblock Matrix completion has no spurious local minimum.
\newblock In {\em Advances in Neural Information Processing Systems}, pages
  2973--2981, 2016.

\bibitem{freeman2016topology}
C~Daniel Freeman and Joan Bruna.
\newblock Topology and geometry of half-rectified network optimization.
\newblock {\em ICLR 2017}, 2017.
\newblock preprint arXiv:1611.01540.

\bibitem{bhojanapalli2016global}
Srinadh Bhojanapalli, Behnam Neyshabur, and Nati Srebro.
\newblock Global optimality of local search for low rank matrix recovery.
\newblock In {\em Advances in Neural Information Processing Systems}, pages
  3873--3881, 2016.

\bibitem{park2016non}
Dohyung Park, Anastasios Kyrillidis, Constantine Carmanis, and Sujay Sanghavi.
\newblock Non-square matrix sensing without spurious local minima via the
  {B}urer-{M}onteiro approach.
\newblock In {\em Artificial Intelligence and Statistics}, pages 65--74, 2017.

\bibitem{du2017gradient}
Simon~S Du, Jason~D Lee, Yuandong Tian, Aarti Singh, and Barnabas Poczos.
\newblock Gradient descent learns one-hidden-layer {CNN}: Don’t be afraid of
  spurious local minima.
\newblock In {\em International Conference on Machine Learning}, pages
  1338--1347, 2018.

\bibitem{ge2017optimization}
Rong Ge and Tengyu Ma.
\newblock On the optimization landscape of tensor decompositions.
\newblock In {\em Advances in Neural Information Processing Systems}, pages
  3653--3663, 2017.

\bibitem{ge2017no}
Rong Ge, Chi Jin, and Yi~Zheng.
\newblock No spurious local minima in nonconvex low rank problems: A unified
  geometric analysis.
\newblock In {\em Proceedings of the 34th International Conference on Machine
  Learning}, pages 1233--1242, 2017.

\bibitem{lu2017depth}
Haihao Lu and Kenji Kawaguchi.
\newblock Depth creates no bad local minima.
\newblock {\em arXiv preprint arXiv:1702.08580}, 2017.

\bibitem{ling2018landscape}
Shuyang Ling, Ruitu Xu, and Afonso~S Bandeira.
\newblock On the landscape of synchronization networks: A perspective from
  nonconvex optimization.
\newblock {\em arXiv preprint arXiv:1809.11083}, 2018.

\bibitem{gross1984simplest}
David~J Gross and Marc M{\'e}zard.
\newblock The simplest spin glass.
\newblock {\em Nuclear Physics B}, 240(4):431--452, 1984.

\bibitem{fyodorov2004complexity}
Yan~V Fyodorov.
\newblock Complexity of random energy landscapes, glass transition, and
  absolute value of the spectral determinant of random matrices.
\newblock {\em Physical review letters}, 92(24):240601, 2004.

\bibitem{auffinger2013random}
Antonio Auffinger, G{\'e}rard~Ben Arous, and Ji{\v{r}}{\'\i} {\v{C}}ern{\'y}.
\newblock Random matrices and complexity of spin glasses.
\newblock {\em Communications on Pure and Applied Mathematics}, 66(2):165--201,
  2013.

\bibitem{sagun2014explorations}
Levent Sagun, V~Ugur Guney, Gerard~Ben Arous, and Yann LeCun.
\newblock Explorations on high dimensional landscapes.
\newblock {\em arXiv preprint arXiv:1412.6615}, 2014.

\bibitem{arous2017landscape}
Gerard Ben~Arous, Song Mei, Andrea Montanari, and Mihai Nica.
\newblock The landscape of the spiked tensor model.
\newblock {\em arXiv preprint arXiv:1711.05424}, 2017.

\bibitem{johnstone2009consistency}
Iain~M Johnstone and Arthur~Yu Lu.
\newblock On consistency and sparsity for principal components analysis in high
  dimensions.
\newblock {\em Journal of the American Statistical Association},
  104(486):682--693, 2009.

\bibitem{deshpande2014information}
Yash Deshpande and Andrea Montanari.
\newblock Information-theoretically optimal sparse {PCA}.
\newblock In {\em Information Theory (ISIT), 2014 IEEE International Symposium
  on}, pages 2197--2201. IEEE, 2014.

\bibitem{richard2014statistical}
Emile Richard and Andrea Montanari.
\newblock A statistical model for tensor {PCA}.
\newblock In {\em Advances in Neural Information Processing Systems}, pages
  2897--2905, 2014.

\bibitem{sarao2019passed}
Stefano Sarao~Mannelli, Florent Krzakala, Pierfrancesco Urbani, and Lenka
  Zdeborova.
\newblock Passed \& spurious: Descent algorithms and local minima in spiked
  matrix-tensor models.
\newblock In {\em International Conference on Machine Learning}, pages
  4333--4342, 2019.

\bibitem{sarao2018marvels}
Stefano Sarao~Mannelli, Giulio Biroli, Chiara Cammarota, Florent Krzakala,
  Pierfrancesco Urbani, and Lenka Zdeborov{\'a}.
\newblock Marvels and pitfalls of the langevin algorithm in noisy
  high-dimensional inference.
\newblock {\em arXiv preprint arXiv:1812.09066}, 2018.

\bibitem{ros2018complex}
Valentina Ros, Gerard~Ben Arous, Giulio Biroli, and Chiara Cammarota.
\newblock Complex energy landscapes in spiked-tensor and simple glassy models:
  Ruggedness, arrangements of local minima, and phase transitions.
\newblock {\em Physical Review X}, 9(1):011003, 2019.

\bibitem{baik2005phase}
Jinho Baik, G{\'e}rard~Ben Arous, Sandrine P{\'e}ch{\'e}, et~al.
\newblock Phase transition of the largest eigenvalue for nonnull complex sample
  covariance matrices.
\newblock {\em The Annals of Probability}, 33(5):1643--1697, 2005.

\bibitem{CHS93}
A~Crisanti, H~Horner, and H-J Sommers.
\newblock The spherical $p$-spin interaction spin-glass model.
\newblock {\em Zeitschrift f{\"u}r Physik B Condensed Matter}, 92(2):257--271,
  1993.

\bibitem{cugliandolo1993analytical}
Leticia~F Cugliandolo and Jorge Kurchan.
\newblock Analytical solution of the off-equilibrium dynamics of a long-range
  spin-glass model.
\newblock {\em Physical Review Letters}, 71(1):173, 1993.

\bibitem{arous2006cugliandolo}
Gerard Ben~Arous, Amir Dembo, and Alice Guionnet.
\newblock {C}ugliandolo-{K}urchan equations for dynamics of spin-glasses.
\newblock {\em Probability theory and related fields}, 136(4):619--660, 2006.

\bibitem{CC05}
Tommaso Castellani and Andrea Cavagna.
\newblock Spin glass theory for pedestrians.
\newblock {\em Journal of Statistical Mechanics: Theory and Experiment},
  2005:P05012, 2005.

\bibitem{crisanti1992sphericalp}
Andrea Crisanti and H-J Sommers.
\newblock The spherical p-spin interaction spin glass model: the statics.
\newblock {\em Zeitschrift f{\"u}r Physik B Condensed Matter}, 87(3):341--354,
  1992.

\bibitem{CK94SK}
Leticia~F Cugliandolo and Jorge Kurchan.
\newblock On the out-of-equilibrium relaxation of the
  {S}herrington-{K}irkpatrick model.
\newblock {\em Journal of Physics A: Mathematical and General}, 27(17):5749,
  1994.

\bibitem{CK95}
Leticia~F Cugliandolo and Jorge Kurchan.
\newblock Weak ergodicity breaking in mean-field spin-glass models.
\newblock {\em Philosophical Magazine B}, 71(4):501--514, 1995.

\bibitem{zamfir2008limiting}
Pompiliu~Manuel Zamfir.
\newblock Limiting dynamics for spherical models of spin glasses with magnetic
  field.
\newblock {\em arXiv preprint arXiv:0806.3519}, 2008.

\bibitem{bouchaud1998out}
Jean-Philippe Bouchaud, Leticia~F Cugliandolo, Jorge Kurchan, and Marc
  M\'{e}zard.
\newblock Out of equilibrium dynamics in spin-glasses and other glassy systems.
\newblock {\em Spin glasses and random fields}, pages 161--223, 1998.

\bibitem{Cu03}
Leticia~F Cugliandolo.
\newblock Course 7: Dynamics of glassy systems.
\newblock In {\em Slow Relaxations and nonequilibrium dynamics in condensed
  matter}, pages 367--521. Springer, 2003.

\bibitem{subag2017complexity}
Eliran Subag.
\newblock The complexity of spherical $ p $-spin models—a second moment
  approach.
\newblock {\em The Annals of Probability}, 45(5):3385--3450, 2017.

\bibitem{adler2009random}
Robert~J Adler and Jonathan~E Taylor.
\newblock {\em Random fields and geometry}.
\newblock Springer Science \& Business Media, 2009.

\bibitem{CL04}
Andrea Crisanti and Luca Leuzzi.
\newblock Spherical 2+ p spin-glass model: An exactly solvable model for glass
  to spin-glass transition.
\newblock {\em Physical review letters}, 93(21):217203, 2004.

\bibitem{arous2018algorithmic}
Gerard~Ben Arous, Reza Gheissari, and Aukosh Jagannath.
\newblock Algorithmic thresholds for tensor {PCA}.
\newblock {\em arXiv preprint arXiv:1808.00921}, 2018.

\bibitem{barrat1997p}
A~Barrat.
\newblock The p-spin spherical spin glass model.
\newblock {\em arXiv preprint cond-mat/9701031}, 1997.

\bibitem{kurchan1996phase}
Jorge Kurchan and Laurent Laloux.
\newblock Phase space geometry and slow dynamics.
\newblock {\em Journal of Physics A: Mathematical and General}, 29(9):1929,
  1996.

\bibitem{krzakala2013performance}
Florent Krzakala and Lenka Zdeborov{\'a}.
\newblock Performance of simulated annealing in p-spin glasses.
\newblock In {\em Journal of Physics: Conference Series}, volume 473, page
  012022. IOP Publishing, 2013.

\bibitem{semerjian2004stochastic}
Guilhem Semerjian, Leticia~F Cugliandolo, and Andrea Montanari.
\newblock On the stochastic dynamics of disordered spin models.
\newblock {\em Journal of statistical physics}, 115(1-2):493--530, 2004.

\end{thebibliography}
\end{document}